\documentclass{article} 
\usepackage{iclr2024_conference,times}

\usepackage{amsmath,amsfonts,bm}

\def\eqref#1{equation~\ref{#1}}

\def\1{\bm{1}}

\DeclareMathAlphabet{\mathsfit}{\encodingdefault}{\sfdefault}{m}{sl}
\SetMathAlphabet{\mathsfit}{bold}{\encodingdefault}{\sfdefault}{bx}{n}

\usepackage{hyperref}
\usepackage{url}

\usepackage{graphicx}
\usepackage{xcolor}

\newcommand{\myparagraph}[1]{\paragraph{#1}\mbox{}\\}

\title{CyberPal.AI: Empowering LLMs with Expert-Driven Cybersecurity Instructions}

\author{Matan Levi$^{1,2}$~, Yair Alluouche$^1$~, Daniel Ohayon$^1$~, Anton Puzanov$^1$ \\
{$^1$\textbf{IBM Research}}~, {$^2$\textbf{Ben-Gurion University}} \\
\texttt{\{matanle,yair,daniel.ohayon,antonp\}@il.ibm.com} 
}

\iclrfinalcopy 
\begin{document}

\maketitle

\begin{abstract}
Large Language Models (LLMs) have significantly advanced natural language processing (NLP), providing versatile capabilities across various applications. However, their application to complex, domain-specific tasks, such as cyber-security, often faces substantial challenges. In this study, we introduce \textit{SecKnowledge} and \textit{CyberPal.AI} to address these challenges and train security-expert LLMs.
\textit{SecKnowledge} is a domain-knowledge-driven cyber-security instruction dataset, meticulously designed using years of accumulated expert knowledge in the domain through a multi-phase generation process. \textit{CyberPal.AI} refers to a family of LLMs fine-tuned using SecKnowledge, aimed at building security-specialized LLMs capable of answering and following complex security-related instructions. 
Additionally, we introduce SecKnowledge-Eval, a comprehensive and diverse cyber-security evaluation benchmark, composed of an extensive set of cyber-security tasks we specifically developed to assess LLMs in the field of cyber-security, along with other publicly available security benchmarks.
Our results show a significant average improvement of up to 24\% over the baseline models, 
underscoring the benefits of our expert-driven instruction dataset generation process.
These findings contribute to the advancement of AI-based cyber-security applications, 
paving the way for security-expert LLMs that can enhance threat-hunting and investigation processes.
\end{abstract}

\section{Introduction}
The rapid progress of LLMs offers a wide range of new capabilities that would have been considered unrealistic only a few years ago. LLMs have emerged as disruptive technology in domains ranging from healthcare to finance, changing the way we consume information and perform our daily tasks. As LLMs are trained on trillions of tokens, they should have fundamental knowledge of most domains available online. 

One such domain is cyber-security. Yet, cyber-security is also a very complex domain. It requires deep understanding in multiple areas of expertise, such as operating systems, network and communication protocols, malware analysis, threat management, and many others. Furthermore, as cyber-security practice spans from security at the physical layer to security at the application layer,
navigating this diverse landscape requires a comprehensive understanding and the ability to connect disparate elements effectively. Therefore, traditional data generation methods will not be effective \citep{mitra2023orca}.
As cyber-security is complex and highly domain-expert-driven, it is required to present LLMs
with domain-specific data generated from expert knowledge to unlock and harness the potential of LLMs in the field.

Over the past decades, security experts have invested considerable time and resources into monitoring cyber activities, investigating incidents, and producing high-quality reports and comprehensive knowledge bases, which include detection rules for identifying and mitigating threats, among other crucial activities. This study seeks to utilize this extensive domain knowledge and, by integrating it with the robust capabilities of LLMs, create a highly valuable instruction-tuning dataset. By combining these efforts, we aim to 
unlock the potential of LLMs in cyber-security. 

To achieve this, we present a domain-knowledge-driven instruction dataset, dubbed \textit{SecKnowledge}.
The objectives of \textit{SecKnowledge} are twofold: firstly, to teach LLMs to follow both simple and complex security instruction better; secondly, to enhance the comprehension of LLMs regarding the wide landspace of cyber-security and the intricate relationships between various security concepts.

Subsequently, we present \textit{CyberPal.AI}, a family of Generative LLMs, fine-tuned from open-source state-of-the-art LLMs, using the \textit{SecKnowledge} dataset. We demonstrate how \textit{CyberPal.AI} outperforms baseline LLMs on various tasks in the cyber-security domain.

Lastly, we construct SecKnowledge-Eval, a diverse and comprehensive set of evaluation datasets
specifically designed 
to test LLMs' ability to develop a holistic understanding of the cyber-security domain and effectively navigate complex security concepts, such as threat investigation. SecKnowledge-Eval consists of evaluation datasets we developed and were designed to assess LLMs on complex cyber-security tasks such as relationships between different security aspects (e.g., the relationship between attack pattern and a specific attack technique), alongside other well-known public cyber-security benchmarks that primarily test general knowledge in the field. 

Overall, we make the following contributions:
\begin{itemize}
    \item We construct \textit{SecKnowledge}, an instruction tuning dataset generated using an expert-driven process on a wide range of security-related datasets. The dataset construction involves two main steps. In the first step, we create instructions based on predefined schemas established through domain expertise. These schemas define templates that are filled with domain-expert knowledge and supplemented with LLM-generated content when necessary. In the second step, we expand the initial dataset through a hybrid synthetic content-based data generation process.
    \item We train \textit{CyberPal.AI}, a family of cyber-security expert LLMs, capable of understanding complex security concepts. \textit{CyberPal.AI} demonstrates the advantages of enhancing LLMs with our domain-knowledge instruction dataset, \textit{SecKnowledge}.
    \item We developed \textit{SecKnowledge-Eval}, a suite of evaluation datasets specifically designed to assess LLMs in the cyber-security domain. \textit{SecKnowledge-Eval} consists of evaluation datasets we constructed to assess LLMs' capabilities on complex cyber-security tasks, alongside public benchmarks, intending to generate a comprehensive and diverse evaluation dataset for assessing both knowledge and understanding of models in the field of cyber-security. CyberPal.AI demonstrated superior performance over its baseline models, showing a substantial average improvement of up to 24\% in training-aligned tasks and up to 10\% in public cyber-security benchmarks.
\end{itemize}

\section{Related Work}
\subsection{General Domains Instruction-Tuning}
Instruction tuning \citep{wei2021finetuned, longpre2023flan, raffel2020exploring, xu2022zeroprompt, sanh2021multitask, chung2022scaling}
demonstrates how fine-tuning Language Models (LMs) with NLP instructions enhances base models' performance in following instructions. In \citep{ouyang2022training} OpenAI trained InstructGPT from GPT-3 \citep{brown2020language}. Vicuna \citep{chiang2023vicuna} used 70k conversations with ChatGPT collected from users via ShareGPT, and used it to fine-tune LLAMA \citep{touvron2023llama}. Other works \citep{sun2024principle, xu2023wizardlm,wang2022self, yin2023dynosaur, xu2023baize} developed various Synthetic Data Generation processes to generate more diverse/complex sets of instructions. In Aplaca \citep{taori2023stanford}, researchers used a small seed of 175 instructions, and prompt ChatGPT to generate a 52k instruction-tuning dataset by utilizing the Self-Instruct \cite{wang2022self} approach. WizardLM \citep{xu2023wizardlm} developed Evol-Instruct that can generate a set of instructions with increasing complexity.

\subsection{Domain specific Instruction-Tuning}
Our work falls within the line of research focuses on developing expert LLMs through instruction tuning in specific domains, such as writing assistants \citep{zhang2023multi}, arithmetic \citep{liu2023goat}, translation \citep{jiao2023parrot}, medicine \citep{thawkar2023xraygpt}, code \citep{codealpaca, luo2023wizardcoder}, and many others.

Specifically for the domain of cyber-security, there have been several works that aimed at training security models.
Although not directly related, a line of works trains Encoder-only architecture on security data either from scratch \citep{bayer2022cysecbert, park2023pretrained} or as continual pre-training \citep{ranade2021cybert, aghaei2022securebert}. However, these models are neither generative nor were trained to follow instructions. For the specific task of fine-tuning generative models for cyber-security applications, VulDetect \citep{omar2023vuldetect} fine-tuned GPT-2 on a dataset containing both vulnerable and non-vulnerable code. The model is fine-tuned to detect anomalies that represent regular behavior. CyberBench was introduced by \citet{liucyberbench} as a cyber-security evaluation dataset that was collected from different works and combined into one security benchmark that includes Name Entity Recognition (NER) tasks for cyber-security corpus, summarization of security blogs, multi-choice Q\&A and Classification tasks.
SecureFalcon \citep{ferrag2023securefalcon} was trained to differentiate between vulnerable and non-vulnerable C code samples, and is specialized in detecting software vulnerabilities.
In contrast to previous efforts, we do not focus on one or more predefined set of tasks. We generate a highly complex and diverse dataset of security instructions spanning a broad spectrum of topics and skills using a domain-expert-driven instruction generation process. As will be described below, we use both domain-expert knowledge alongside LLM generation capabilities to populate our security instruction dataset. This comprehensive dataset enables us to train general-purpose security models.

\section{SecKnowledge: Domain-knowledge driven Cyber-security Instruction dataset}
This section details the construction of \textit{SecKnowledge}, a novel instruction tuning dataset specifically designed for the domain of cyber-security. We leverage expert knowledge and employ a two-step process to build a comprehensive and diverse dataset capable of supporting instruction tuning for various security-related tasks. The two-step process is defined as follows:

\begin{enumerate}
  \item The first generation step focuses on creating high-quality instructions based on predefined schemas. These schemas are established through experts-driven in-depth analysis of the diverse set of security datasets, their individual characteristics, and the relationships between different entities within and between datasets. This ensures that the instructions are relevant, accurate, and capture the nuances of various security concepts and tasks.
More specifically, each predefined schema consists of rules by which the data-source should be processed into instructions using parsers we developed, ensuring  that the generated instructions focus on the important and unique characteristics of the data-source, and are representative of real-world security scenarios. Our method can be considered as an extension to methods such \cite{wei2021finetuned, longpre2023flan}, where templates are simply assigned with predefined questions and answers. In Section \ref{domain-kwonledge-dg} we break down the generation process.
  \item The second generation step expands the generated initial dataset and improves its diversity and complexity. To do so, we employ a hybrid synthetic content-grounded data generation process. 
More specifically, we fused Evol-Instruct \citep{xu2023wizardlm} and Self-Instruct \citep{wang2022self} and combined them with content-grounded generation and evaluation pipelines. Additionally, we implemented a routing mechanism between the two generation methods that helps to reduce hallucinations.
This process leverages the initial set of instructions and data from the first generation step to generate additional instructions that follow the established schemas but increase the model's overall generalizability. By incorporating content-grounded synthetic data, we increase the diversity and volume of the final dataset, ultimately leading to more robust and capable security models. In section \ref{SDG}, we further elaborate on the specifics of the generation process.
\end{enumerate}

Our final SecKnowledge dataset consists of various instruction types, among which are: open/closed book question answering, yes/no questions, multi-choice Q\&A, Chain of Thoughts (CoT) \citep{wei2022chain}, summarization, logic validation, odd/leave one out multi-choice Q\&A, question generation, query/rule explanation and generation, TTP mapping, and others.

Table \ref{table:ds-overview} summarizes the security instruction sets composed in the first generation step.
Unless otherwise specified, we use the open-source Mixtral \citep{jiang2024mixtral} model for both data generation and evaluation processes.

\subsection{First Generation Step: Domain knowledge-driven instruction generation}
\label{domain-kwonledge-dg}
Leveraging domain expertise, we first parse and enrich each one of the various security data sources using their unique characteristics and structure, derive connections between the  documents in each data-source, and even derive connections between different data sources, as we will describe in the upcoming sections. 

We establish a set of predefined, domain-knowledge-driven, schemas that capture the essential elements of different security tasks. Each schema consists of a series of pre-defined, domain-expertise-driven rules.
Each rule is then translated into a parsing object. The parsing object will then generate and fill the instruction templates with the parsed data.

These schemas enable building instructions that capture each dataset's unique objectives and characteristics. This approach ensures that the instructions accurately reflect the desired model behavior and provide a strong foundation for effective instruction tuning.

The subsequent paragraphs provide a detailed description of the data sources and methodologies employed in the first step of the SecKnowledge data generation process.

\begin{table}[h]
{\small
\centering
\begin{tabular}{l|r}
Dataset                  & \# of generated instruction \\ \hline
MITRE ATT\&CK            & 45,901                     \\
CWE                      & 4,080                      \\
CVE                      & 8,447                      \\
CAPEC                    & 3,917                      \\
Security Wiki            & 11,000                     \\
Security interview Q\&A  & 500                        \\
Threat reports           & 4,500                      \\
BRON                     & 62,227                     \\
SIEM alert TTP mapping   & 400                        \\
Sigma rules              & 9,329                      \\
Security Stack Exchange  & 2,573                      \\ \hline
Total                    & 152,874                  
\end{tabular}
\caption{Overview of the initial instructions constructed from the  datasets on the \textbf{first step} as described in \ref{domain-kwonledge-dg}. These instructions will be used as the \textbf{seed for the second generation step}.}
\label{table:ds-overview}}
\end{table}

\subsubsection{Structure-driven instruction generation} 
\label{structure-section}
The straightforward method for creating instructions 
dataset from the documents is to provide a teacher model with raw documents and instruct it to generate instructions based on the content.
However, this method presents several challenges. 
Relying on models to produce instructions that simultaneously capture the unique characteristics of datasets while maintaining complexity and diversity proves to be a difficult task.
One reason is that models tend to focus on specific or localized sections of a document when generating instructions. 
More significantly, models struggle to capture and exploit the relationships between different components within each dataset and the relationships between different datasets.


In this section, we introduce a different method for generating an instruction set from documents. Our approach exploits the structured nature of the various cyber-security documents to create a high-quality, diverse, and complex instruction dataset. We demonstrate the efficiency of our method using the MITRE framework, a comprehensive security resource that encapsulates years of expertise in the security domain. We developed specialized parsers that use predefined schemas and rules that harness the structured nature of the data 
to generate instructions. These parsers extract relationships between different entities within the datasets and, using their corresponding schemas, transform these documents into instructions

More specifically, we utilize the structured nature of the different MITRE datasets, among which we can find:
\begin{itemize}
\item MITRE ATT\&CK - comprehensive knowledge base of adversary tactics and techniques based on real-world observations, MITRE ATT\&CK also provides detection and mitigation methods for each technique and their use by threat groups and software tools\footnote{https://attack.mitre.org}.
\item CWE (Common Weakness Enumeration) - community-developed list of software and hardware weakness types, serving as a common language for describing security vulnerabilities\footnote{https://cwe.mitre.org}.
\item CVE (Common Vulnerabilities and Exposures) - dictionary of publicly known cyber-security vulnerabilities and exposures, aims at
standardizing the way vulnerability information is shared\footnote{https://cve.mitre.org}. 
\item CAPEC (Common Attack Pattern Enumeration and Classification) - a structured catalog of common attack patterns describing how adversaries exploit application weaknesses and other cyber-enabled capabilities. Attack patterns describe the common attributes and approaches that adversaries employ to exploit known weaknesses in cyber-enabled capabilities\footnote{https://capec.mitre.org}.
\end{itemize}
All of which are oriented towards the domain of cyber-security. The compilation of these frameworks encompasses a vast repository of cyber-security domain knowledge and offers extensive coverage of the security field, making it an excellent resource for fine-tuning our model to the specific requirements and nuances of cyber-security. For additional information on each framework, see appendix \ref{mitre_data_appendix}.

Each MITRE framework comprises a structured format that categorizes different aspects of the subject matter, enabling organized analysis of the different security aspects.
As such, we create a schema for each framework. The schema results in the following types of instructions:
\begin{enumerate}
    \item A set of instructions designed to teach the model the specific characteristics of each object (i.e., tactic, technique, mitigation, detection, attack pattern, etc.) For instance, an instruction could detail the relationships between an attack pattern and its corresponding severity, prerequisites, or consequences.
    \item A set of instructions designed to guide the model in understanding the relationships between different objects \textit{within} each dataset.
\end{enumerate}

\begin{figure}[h]
  \centering
  \includegraphics[width=0.65\textwidth]{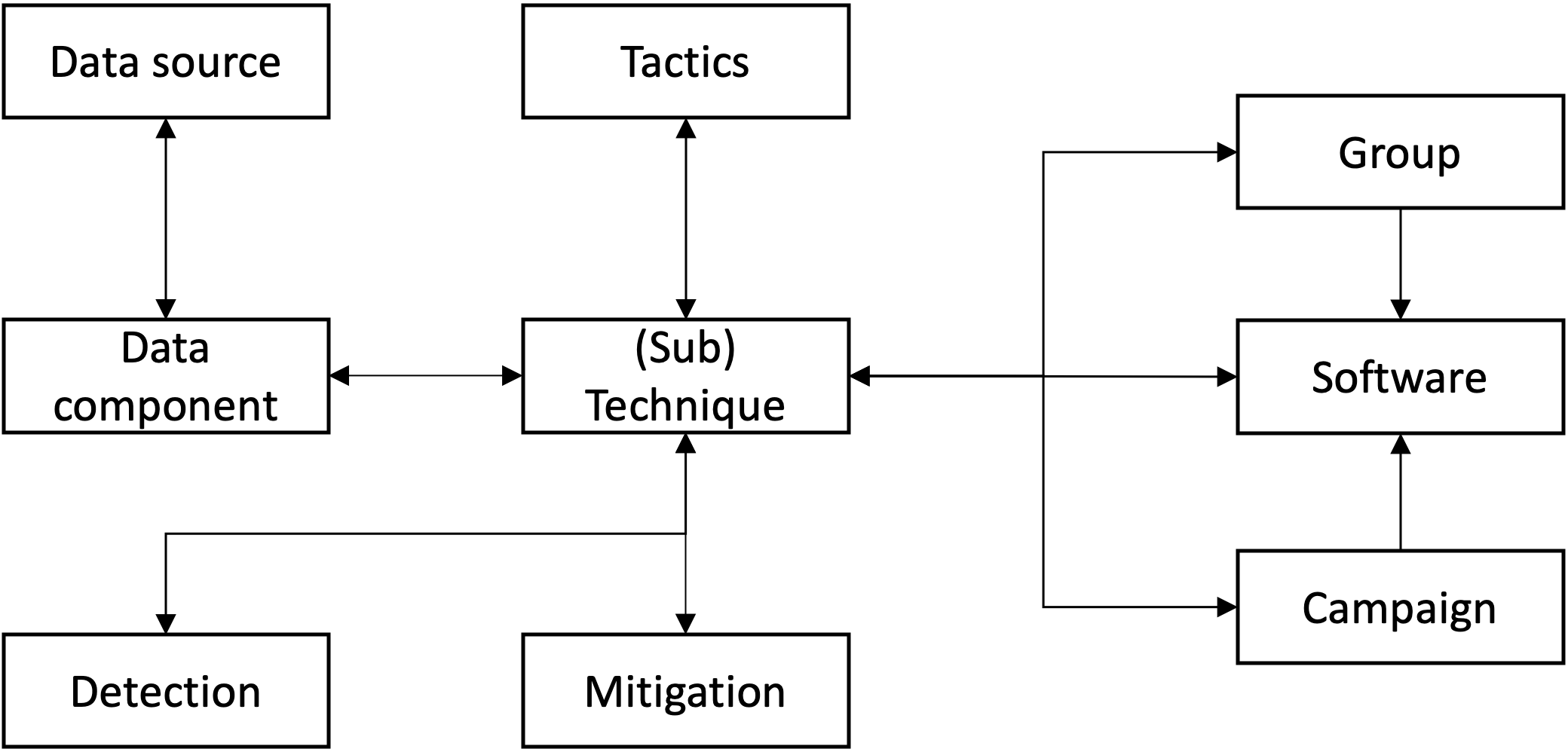}
  \caption{Relationship between different MITRE ATT\&CK components.}
  \label{figures:mitre-fig}
\end{figure}

\begin{figure}[h]
  \centering
  \includegraphics[width=1\textwidth]{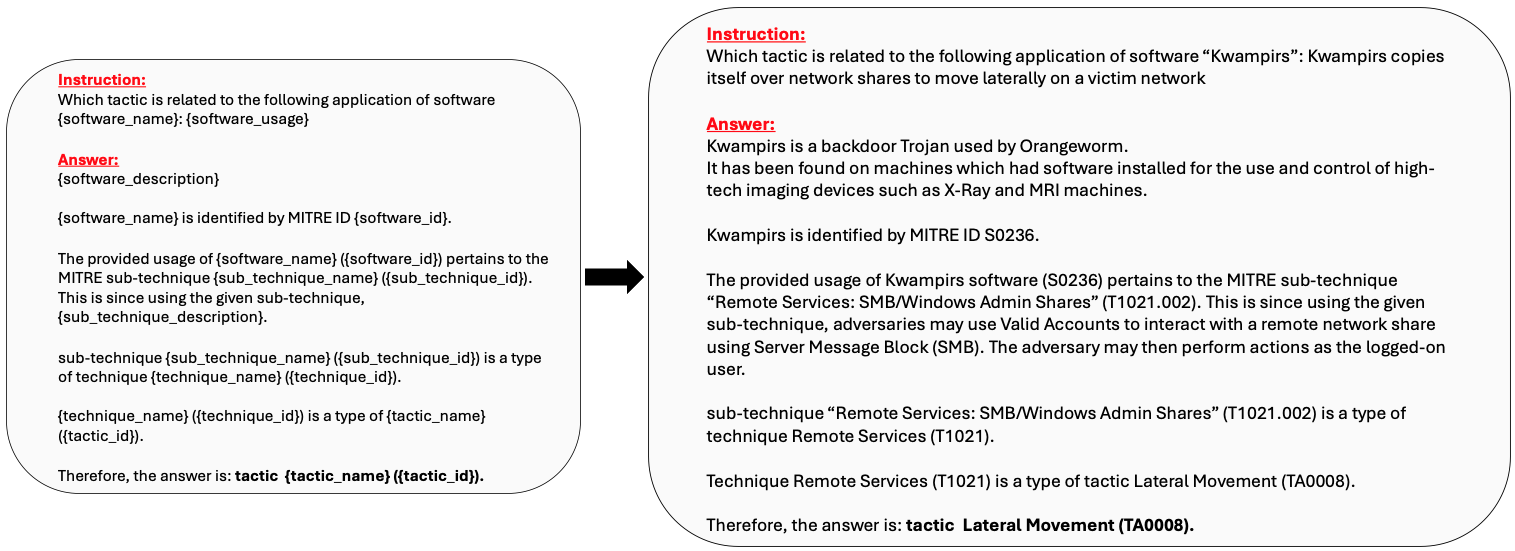}
  \caption{Example of constructing CoT by utilizing our knowledge of the structure relationships between different components within the MITRE ATT\&CK framework. On the left side, there is a template to map from a given malware usage to its corresponding tactic. On the right side, the template is assigned with a specific malware usage and its chain of connections up to the relevant tactic.}
  \label{fig:cot-example}
\end{figure}

Next, we provide an example of chain-of-thoughts instruction generated by utilizing the MITRE ATT\&CK structure.
Figure \ref{figures:mitre-fig} demonstrates the relationships between different objects within the MITRE ATT\&CK framework. Using these relationships, complex instructions are constructed on the wide range of the attack land-space. For instance, see Figure \ref{fig:cot-example}, where Chain of Thoughts (CoT) training example is created. This process begins by asking about the usage of specific malicious software and identifying the tactic to which this software is related. Our parser then relies on the knowledge of the relationships between different components (tactics, techniques, sub-techniques, software) to derive the chain of connections from the specific malicious software to the relevant exploited tactic.

Note that no language model is used during the construction; the connections and relevant text are derived based on our knowledge of the dataset’s structure.

\subsubsection{Structured LLM-Augmented Instruction Generation} 
In the previous section, we demonstrated how to use raw data alongside domain expertise to populate our schema templates. Here, we combine the same abilities of robust predefined schemas and domain knowledge with the flexibility and reasoning abilities of LLMs to create comprehensive instructions.
This approach leverages structured templates for consistency while utilizing a teacher model to dynamically generate and fill specific content, ensuring both accuracy and adaptability in instruction creation. More specifically, our main goal of using the teacher model is not to generate general content, but rather to harness the reasoning capabilities of the teacher model to guide our models to reason on complex security concepts based on the information we provide.

We use structured LLM-augmented instruction generation on the following datasets: BRON, SIEM Rules to TTP Mapping, and Sigma Rules.

\paragraph{BRON:} BRON \citep{hemberg2021linking} is a graph that interconnects threat data sourced from MITRE ATT\&CK, CAPEC, CWE, CVE, MITRE Engage, MITRE D3FEND, MITRE CAR, and exploitdb. This interconnected graph enhances the capabilities of security researchers in conducting advanced threat hunting.

\begin{figure}[h]
  \centering
  \includegraphics[width=0.6\textwidth]{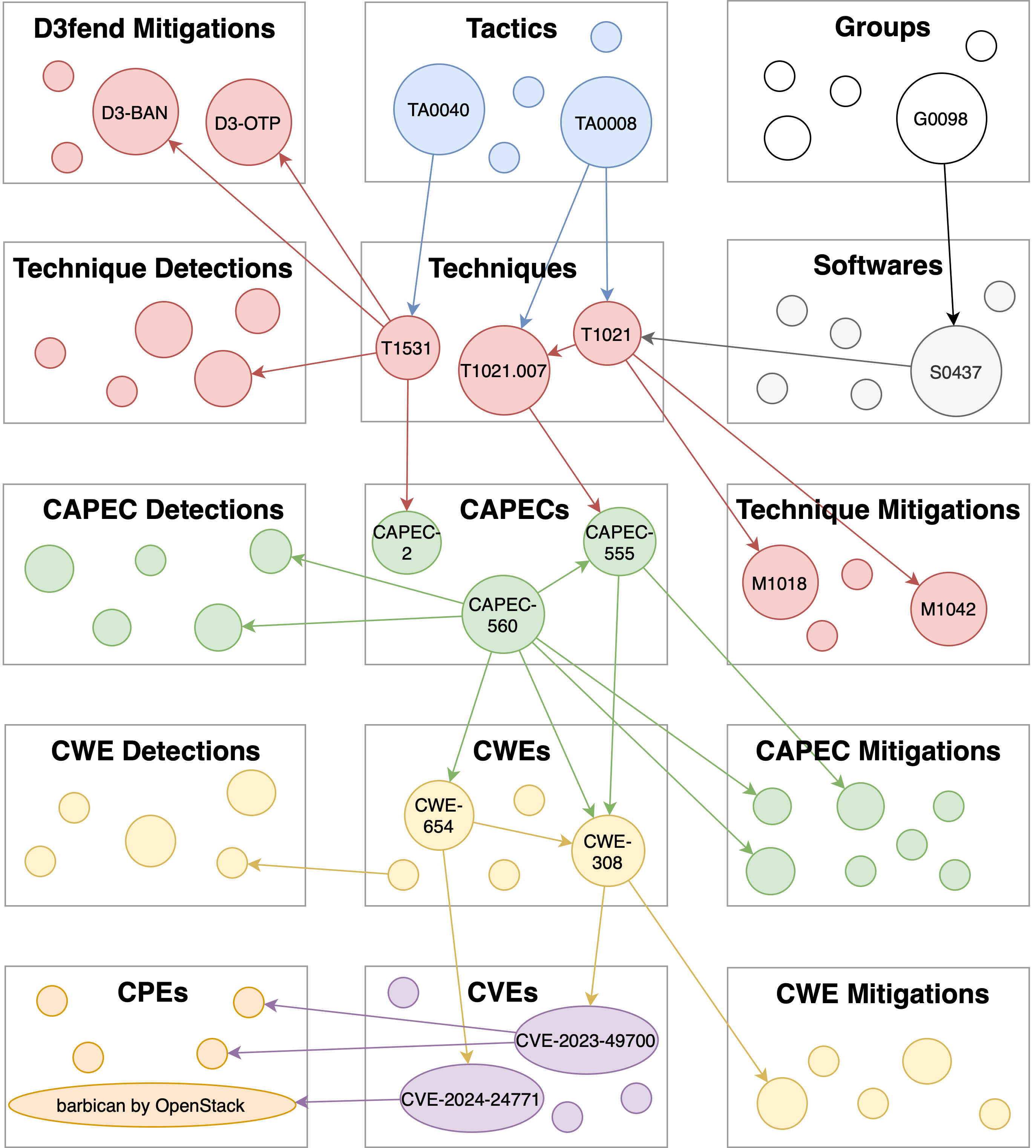}
  \caption{BRON high-level graph structure overview}
  \label{fig:bron}
\end{figure}

After demonstrating in section \ref{structure-section} how the MITRE frameworks can be utilized (individually) to generate instructions on the specific characteristics of each MITRE object and the relationships between different objects \textit{within} each framework, we will leverage BRON to generate instructions on the relationships between different objects \textit{across} frameworks.
See Figure \ref{fig:bron} for a high-level overview of the graph’s structure.

With hundreds of thousands of nodes and millions of edges interconnecting them, BRON's sheer scale makes it impractical to feed directly to an LLM with the expectation of comprehensively learning all relationships. 
Therefore, our primary objective is to generate an instruction set that teaches the model to \textit{reason} if and how different entities are related.



Specifically, using BRON, we have two main goals:
1) construct instructions that will guide LLMs how to reason if two consecutive entities are related to each other (e.g., CWE and CVE nodes), and 2) showcase the reasoning process for LLMs to derive the path from a specific entity of interest to any other entity in the graph
, to accommodate user instruction. This reasoning process will 
enable a more comprehensive understanding of the 
relationships between different entities, such as the connection between a platform and its relevant weaknesses, which are not directly related.

To meet the stated goals, the graph should be processed and traversed into paths, which we will later enrich with domain knowledge from different resources. These paths and explanations will effectively become chain-of-thoughts examples that can guide LLMs to perform effective, complex threat-hunting reasoning.
Our processing requires four steps: path extraction - in which we perform a walk on the graph and extract paths, deriving connections between direct nodes, constructing CoT instructions based on the extracted paths, and multi-path CoT.

\subparagraph{Paths Extraction}First, we gather all one-step paths between nodes of different types that are directly connected in the graph (e.g., all connections between tactic nodes and technique nodes). Next, for all non-direct paths, we perform a random walk on the graph and construct up to 5000 paths between each pair of node types that are not directly connected (e.g., paths between tactic nodes and CVE nodes).

\subparagraph{Derive the Connection Between Direct Nodes}
After extracting one-step paths between nodes of different types (e.g., CAPEC node and CWE node) that are directly connected in the graph, we take these direct links and use the reasoning process of a teacher model to explain the connection between each pair of nodes. 
This involves sending the teacher model instructions that include the descriptions, alongside other information about each node, for each pair of nodes, requiring the teacher model to examine the information and decide if and how the nodes are connected. 
Additionally, we incorporate negative sampling to illustrate that not all nodes in the graph are connected, compelling the model to make decisions based on the nodes’ information.
The negative sampling stage is pivotal as it's impractical to present all existing paths (amounting to millions) to the models we fine-tune. Instead, as our models already see these data sources (ATT\&CK, CWE, etc.) separately during the fine-tuning process, we aim to equip them with the ability to ascertain whether two nodes are linked based on their information, in the expectation that our fine-tuned models will generalize the reasoning process to paths they haven't encountered during training. 

\subparagraph{Constructing CoT on Paths}Additionally, we present the model with longer paths that involve multiple nodes, between non-direct nodes. When constructing these paths, we need to generate an explanation for the connection between each two nodes in the path. To generate the explanations, we enrich the paths with the information we have on each node from our other datasets (MITRE ATT\&CK, CAPEC, etc.). More specifically, for each edge in the path, we use the teacher model to explain the connection between the nodes based on the relationship status (e.g., CVE is a specific implementation of CWE) and the nodes' information (e.g., description) we provide, similar to the direct node processing. This explanation is then attached to the edge between the two nodes. As a result, a Chain-of-Thought (CoT) explanation is generated for each path. 
  
\begin{figure}[h]
  \centering
  \includegraphics[width=1\textwidth]{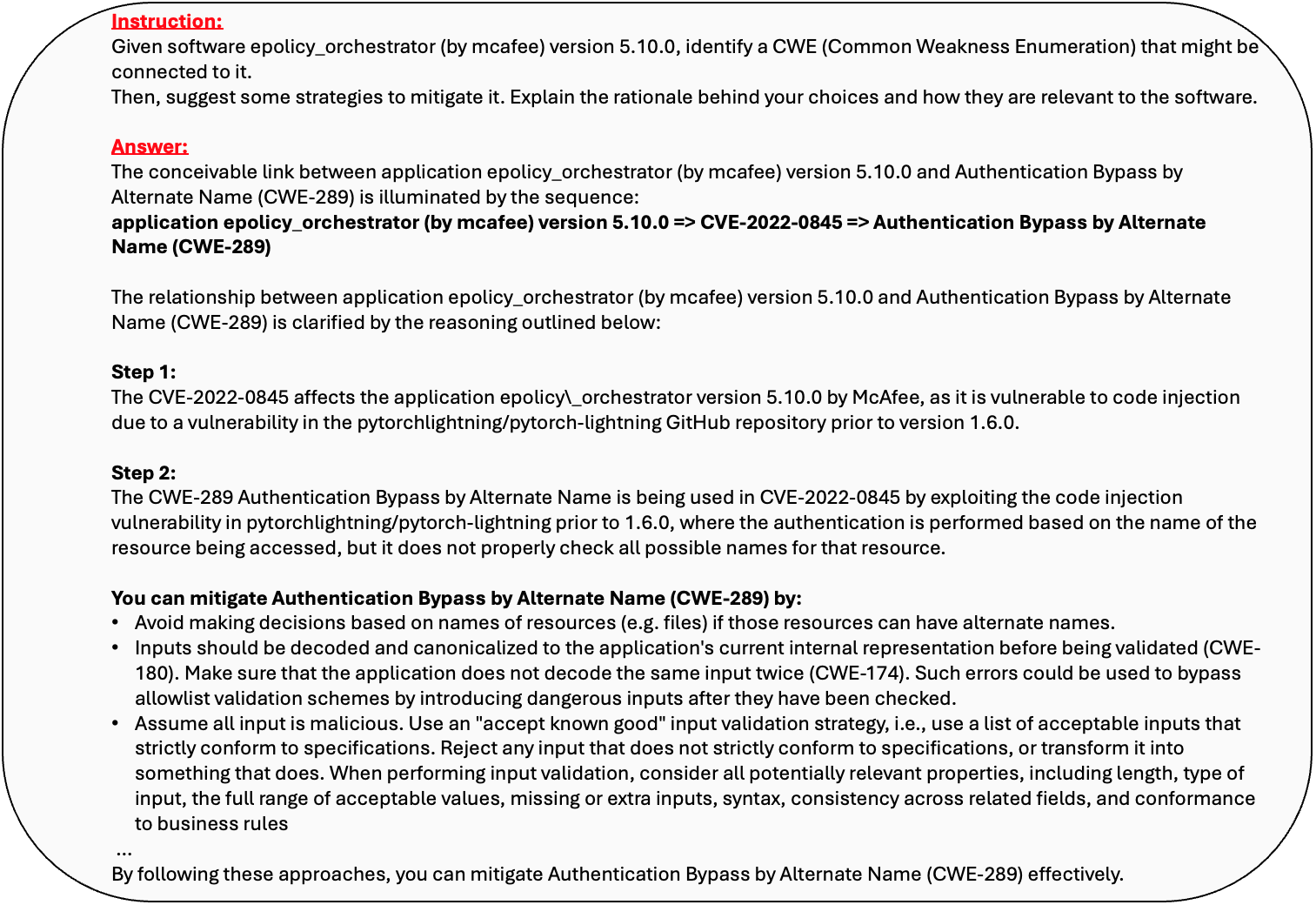}
  \caption{Illustration depicting the construction of a CoT by extracting paths from BRON, enriching them with data and domain knowledge, and using LLM to formulate connections based on the provided information.}
  \label{fig:bron-example}
\end{figure}

\subparagraph{Multi-path CoT}Lastly, we construct more complex instructions, i.e., instructions that can be answered only by involving multiple paths from the graph. One example can be a two-stage/paths instruction where we can first ask what the relevant attack patterns for a given weakness are, and in the second stage how to detect/mitigate them. See Appendix \ref{bron_appendix} for additional details and Figure \ref{fig:bron-example} for an example.

\paragraph{SIEM Rules to TTP Mapping:}
SIEM (Security Information and Event Management) is a security platform that monitors and correlates threat intelligence, network, and user behavior anomalies to prioritize high-fidelity alerts. We have collected a list of 400 rules from IBM's SIEM, QRadar, along with their corresponding Tactics, Techniques, and Procedures (TTP) mappings.

TTP mapping of detection rules is critical in cyber-security as it enables organizations to systematically identify and counteract specific adversary behaviors, thereby enhancing the precision and effectiveness of threat detection.

QRadar's rules are well-structured and include fields such as rule ID, description, pattern, relevant MITRE tactic/technique ID and name, rule risk level, and more. In the following, we will demonstrate how we leverage this structure to develop a series of instructions for educating CyberPal.AI on mapping rules to Tactics, Techniques, and Procedures (TTPs). Our goal is not merely to create a simple mapping task but to teach the model to reason about TTP mapping. To achieve this, we combine expert knowledge with LLMs to generate a comprehensive TTP reasoning instruction dataset, as we describe below.

\begin{figure}[h]
  \centering
  \includegraphics[width=1\textwidth]{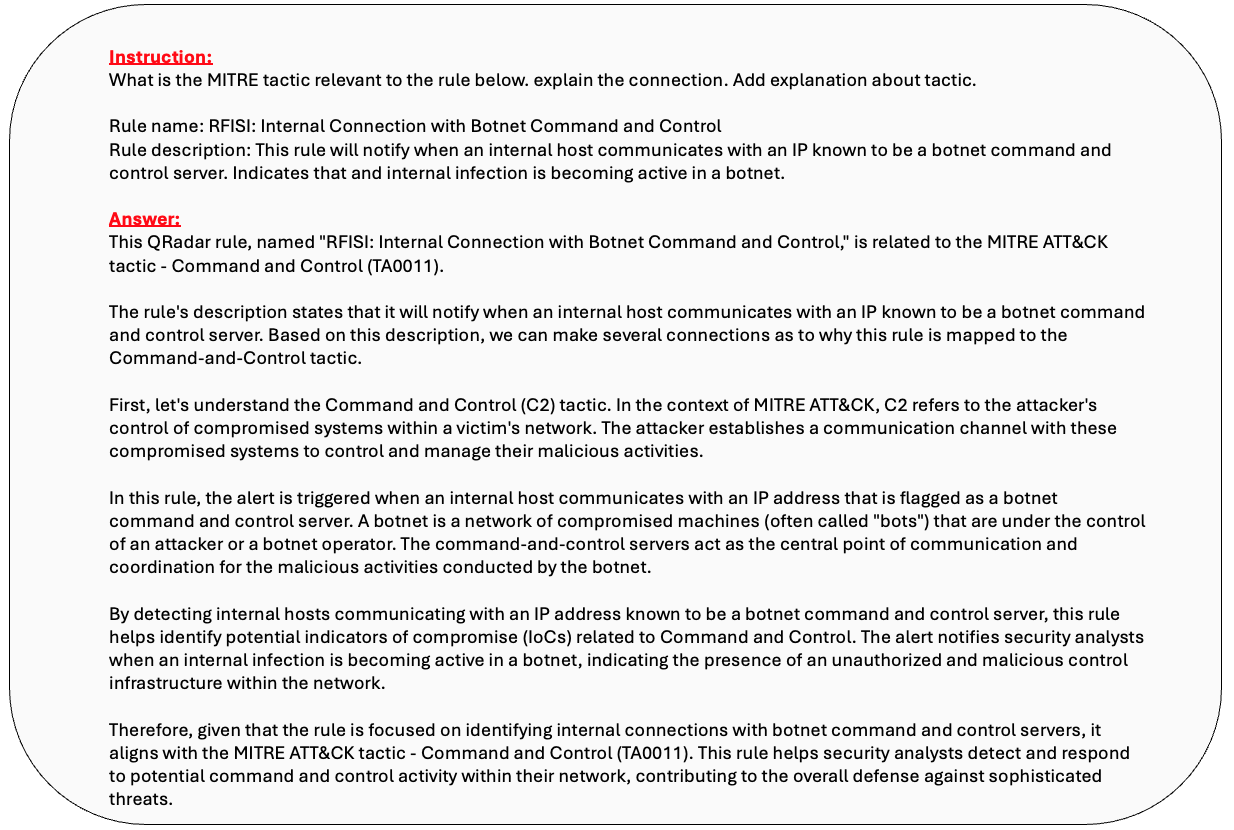}
  \caption{Example of generated instruction using our mapping process from SIEM rules to TTPs. The answer is the explanation that was generated in our construction process.}
  \label{fig:siem}
\end{figure}

The process of creating the TTP mapping instruction set involves retrieving the rule description, tactic/technique ID, and name for each rule and its corresponding TTP mapping. Using this information, we access the description and additional relevant data of the tactic/technique from the MITRE ATT\&CK framework. We tailor a specific schema, that leverages the required information, and guides the teacher model to reason and clarify the relationship between the rule and the provided tactic/technique based on their descriptions and additional relevant data. Subsequently, the model generates an explanation, which undergoes evaluation for correctness by another teacher model (evaluator). Upon acceptance by the evaluator, a set of instructions is formulated based on the rule, the TTP mapping, and the explanation.
The goal of the generated set is to demonstrate to \textit{CyberPal.AI} the reasoning approach of mapping between rules and TTPs.
See Figure \ref{fig:siem} for an example of such instruction.

\paragraph{Sigma Rules:} Sigma is a structured and open signature format that allows to define and describe detection logic. The rule format is flexible and platform-agnostic. The main purpose of Sigma is to provide a structured form in which researchers and analysts can describe their developed detection methods and make them shareable. 
SigmaHQ\footnote{https://github.com/SigmaHQ/sigma} is the main rule repository where detection engineers, threat hunters, and all defensive security practitioners collaborate on detection rules. Here, we leverage the repository's dataset, which contains over 3000 diverse and reliable detection rules as a baseline for our rule instruction set.

Sigma rules contain multiple fields, among which are: the "logsource" field which specifies the type of log data the rule applies to, and the "detection" field which defines the specific conditions that trigger the rule, including event attributes, expected values, and filters for accurate detection. The "level" field indicates the severity of the detected event. Each Sigma rule is connected to the attack it tried to detect.

We take advantage of the Sigma rules structure, and feed the relevant fields to a teacher model, 
and construct the following types of reasoning instructions:
\begin{itemize}
    \item {
    Step-by-step attack detection explanation using log source and rule detection filters within the detection field
    }
    \item {Step-by-step reasoning for attack type mapping via detection indicators}
    \item {Sigma rule generation from attack type and/or detection indicators}
\end{itemize}

\begin{figure}[h]
  \centering
  \includegraphics[width=1\textwidth]{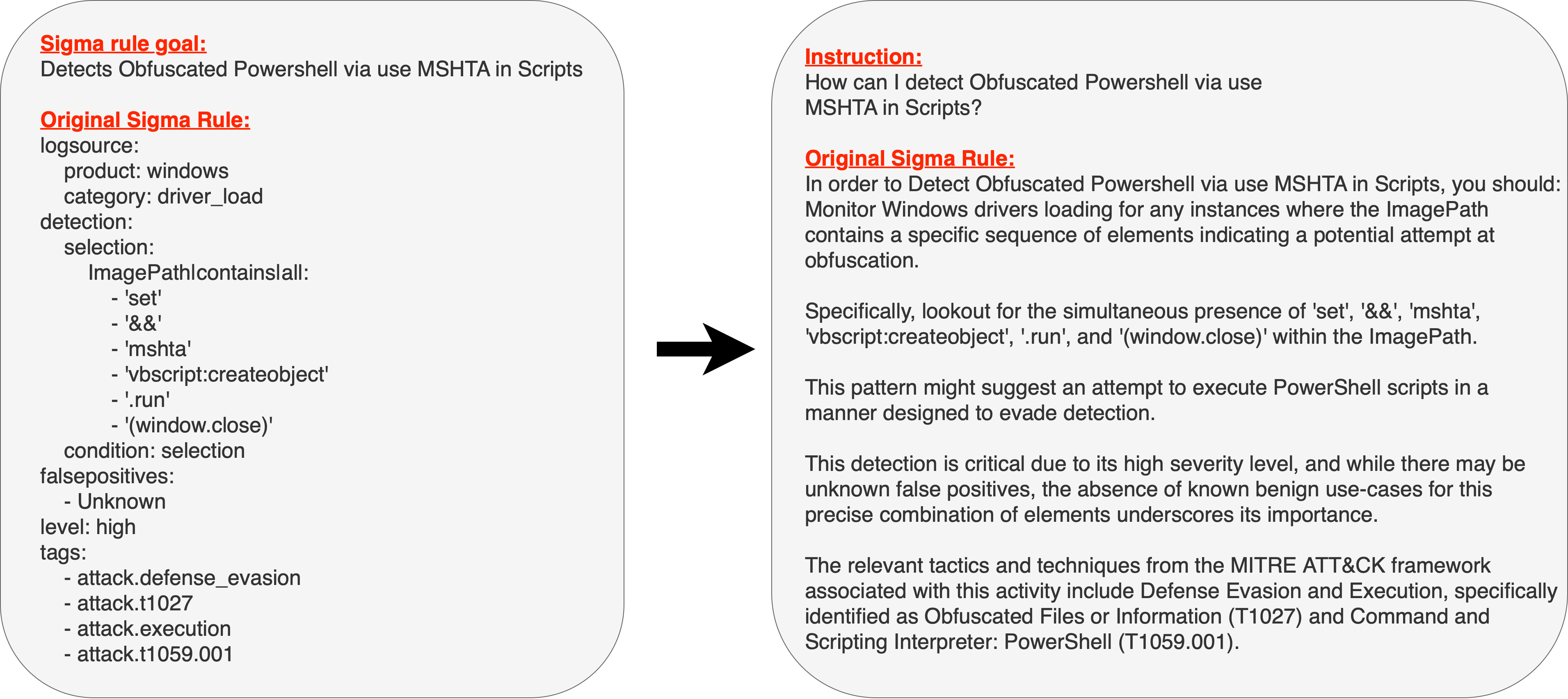}
  \caption{Sigma instruction example - from the original Sigma rule, we construct a task to explain how to detect a specific attack. The Sigma rule will be processed, and based on the relevant instruction template and model prompt, we will construct the instruction and its corresponding answer.}
  \label{fig:sigma}
\end{figure}

We customize a schema for each task type, ensuring it contains the necessary information. See Figure \ref{fig:sigma} for an example of such instruction.
As with other generation processes, we also apply an evaluator (LLM) that tests the correctness of the generated text.

\subsubsection{Additional datasets}  We have collected and generated security-related instructions on various other security datasets: security interview Q\&A, threat reports on various security threats, security and reverse-engineering stack-exchange, and Wikipedia pages related to computer security. For each dataset, we define a schema based on its structure and build instructions in a similar manner to the previously mentioned datasets.


\subsection{Second Generation Step: Content-Ground Synthetic data generation}
\label{SDG}
In the second step of our security instruction generation process, we expand the generated initial dataset from \ref{domain-kwonledge-dg} and improve its diversity and complexity.

For the purpose of synthetic data generation, we build upon the ideas of Self-Instruct \citep{wang2022self} and Evol-Instruct \citep{xu2023wizardlm} and fuse them alongside content-grounded generation and an instruction routing mechanism.

Evol-Instruct starts with an initial set of instructions and rewrites them step by step into more complex instructions. More specifically, Evol-Instruct uses In-depth and In-breath evolving. The In-depth Evolving includes five types of operations: adding constraints, deepening, concretizing, increasing reasoning steps, and complicating input. The In-breadth Evolving is a mutation, i.e., generating a completely new instruction based on the given instruction.

Self-Instruct also starts with an initial set of instructions. Then, the model is prompted to generate instructions for new tasks. It leverages the existing collection of instructions to create more broad-coverage instructions that define, often new, tasks. We can think of In-breadth Evolving as a variant of Self-Instruct.

In the cyber-security domain, accuracy is paramount, as hallucinations can lead to disastrous consequences, such as applying incorrect detection and mitigation strategies, resulting in security breaches and inadequate organizational responses.

Therefore, we fuse the two methods with content-grounded generation, meaning we try to force the model to generate a more complex instruction that is grounded by the document from which the previous instruction was generated.

\begin{figure}[h]
  \centering
  \includegraphics[width=0.7\textwidth]{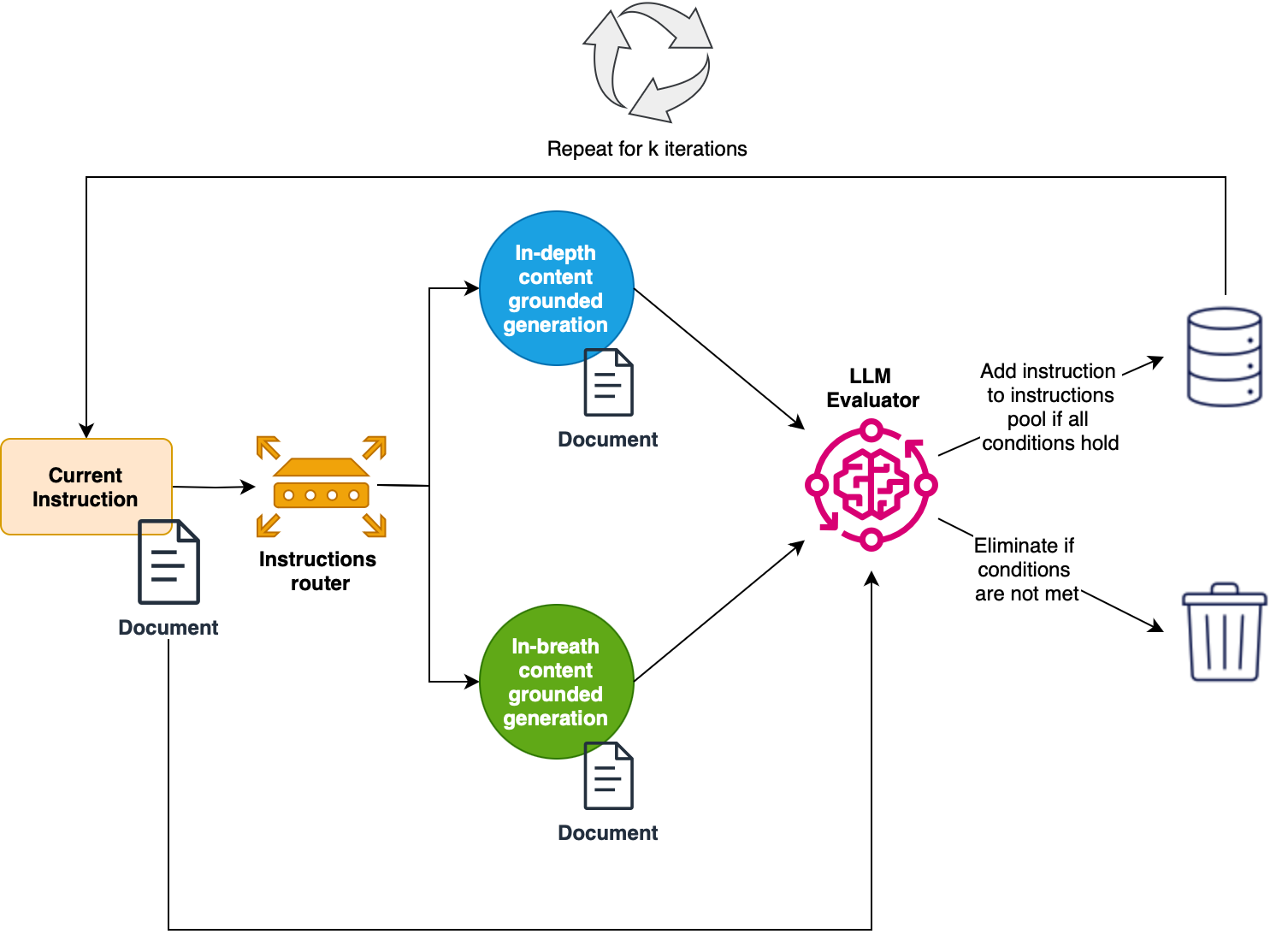}
  \caption{High-level illustration of the content-grounded Synthetic Data Generation (SDG) process.}
  \label{fig:sdg}
\end{figure}

We found that using Evol-Instruct with content-grounded generation tends to diverge and generate inaccurate instructions after several iterations (usually around 3 iterations), resulting in instructions that an LLM cannot answer, non-grounded instructions or instructions that deviate from the relevant topic. 
Therefore, we incorporate a dynamic mechanism that combines Evol-Instruct together with Self-Instruct (See the "Instructions router" in Figure \ref{fig:sdg}), where in the early stages of the synthetic data generation, we mainly focus on In-depth Evolving with the goal of generating more complex instructions on the same topic, and as the generation process progresses, we shift the focus towards Self-Instruct (which can be thought of as 
equivalent to the In-breath instruction generation from Evol-Instruct), where we mainly focus on generating new tasks, while keeping them grounded and in the same domain as the document. More specifically, the probability of Self-Instruct being chosen is doubled every two iterations.
We find that combining Evol and Self-Instruct leads to better content-grounded instructions (due to the difficulty of preserving complex content-grounded instructions in later stages). 

Additionally, we incorporate an internal evaluation mechanism using an LLM evaluator. The evaluator is defined by the following objectives:
\begin{itemize}
    \item {Evaluate if the new instruction is more challenging/complex/rare or diverse}
    \item {Evaluate if the new instruction is of the same domain as the given instruction based on the document, and evaluate that new instruction can be answered by the document}
    \item {Evaluate if  the generated answer correctly answers the new instruction, and that the generated answer is grounded by the document}
\end{itemize}

An instruction will be added to the instruction pool and be used in the next iteration only if it passes the evaluator's assessment. 
Note that not all datasets are suitable for full SDG (i.e., synthetic data generation without domain expert intervention).
For example - we cannot use SDG without domain expert intervention on BRON as it may hallucinate paths. We also cannot use SDG to generate new rules as it is not trivial for a model to generate new and accurate rules in non-common rule languages (e.g., in Sigma rules).
Therefore, we employ our content-grounded SDG process to Security Wiki, Security interview Q\&A, Security Stack Exchange, MITRE ATT\&CK, CWE, CVE, and CAPEC. In the second generation step, we generate an additional 250,000 instructions, consisting of all instructions from different iterations stored in the instruction pool. In total, the final dataset comprises approximately 400,000 complex and diverse instructions following the two generation steps.

\section{SecKnowledge-Eval: A Comprehensive Security Evaluation Dataset}
To assess CyberPal.AI's performance, we constructed a diverse set of seven new evaluation datasets aimed at testing the model's capabilities in cyber threat intelligence. 
To ensure no data contamination between the fine-tuning and testing phases, we partitioned the raw documents into train and test sets, such that the model did not encounter any test-related documents during fine-tuning. For example, if a specific CWE instance was included in the test split, it was not seen by the model during the fine-tuning process. 
After splitting the data, we transformed the documents from the test split into the evaluation tasks described below.
Furthermore, we benchmarked CyberPal.AI against another seven public and general cyber-security evaluation datasets to demonstrate its robustness and comprehensive understanding of security concepts. Overall, our evaluation benchmark consists of 14 diverse datasets, with various task types.
To the best of our knowledge, this is the most comprehensive cyber-security evaluation benchmark, composed of an extensive set of cyber-security tasks alongside other publicly available security benchmarks. For high-level statistics about SecKnowledge-Eval, see Appendix \ref{eval-stat}.
The following paragraphs introduce these tasks and datasets.


\myparagraph{Multiple Choice Tasks} 
Questions in this section are formatted with four multiple-choice answers, similar to the question template from \citep{hendrycks2020measuring}.

\setlength{\parindent}{2em}\textbf{Adversarial MITRE ATT\&CK}
We compiled this dataset using various MITRE ATT\&CK sources. It is designed to assess the model’s knowledge of malicious software, campaigns, attack tactics and techniques, data sources, and potential detections and mitigations for different attacks. The input consists of information about a given MITRE instance (e.g., description), and the correct answer is the source from which it was derived.

To enhance the dataset's difficulty and test our models' robustness, we developed a novel adversarial attack \citep{goodfellow2014explaining, carlini2017adversarial, levi2023splitting} for multi-choice questions on closed domains. For example - in domains where the options for the answer are taken from a closed list of possible options, our goal is to choose the false options that will confuse the model with the highest probability without manipulating the question itself. The main idea is as follows: assume a multiple-choice question, where the choices are taken from a closed list of possible options, with size \textit{k}. For each of the \textit{k-1} false options, we construct a new classification question, where the question is the original question, and there are two options the model should select from: one is the correct option and the other is one of the \textit{k-1} false options. For each such question, we query an LLM with \textit{k-1}
such queries, and select the false options that were the most likely to be selected with respect to the question and the correct option.

As an example, given an instruction \textit{I} from a specific domain within the MITRE framework such as techniques, we generate \textit{k-1} queries for each \textit{I}, where \textit{k} represents the number of all existing techniques, and subsequently query a third-party LLM \textit{k-1} times in the following way: for each false technique option from the list of \textit{k} techniques, the LLM is tasked to classify whether the instruction corresponds to the correct technique or the false technique. When constructing the adversarial questions, our objective is to identify techniques where the Log-Likelihood is the lowest when compared to the correct answer.
Appendix \ref{adv_mc_appendix} presents more details about the attack, which results in our adversarial evaluation dataset.

\setlength{\parindent}{2em}\textbf{SIEM Rule TTP Mapping}
SIEM solutions usually include rules that detect a wide range of activities, including excessive firewall denies, multiple failed login attempts, and potential botnet activity.
We developed a dataset comprising IBM's QRadar rules, aiming to classify each rule according to the appropriate tactic or technique.
This TTP classification dataset is structured as a multiple-choice dataset with four options due to multiple possible correct tags.

\setlength{\parindent}{2em}\textbf{CTI Detection and Mitigation Mapping}
As outlined, BRON captures the interrelationships between different Cyber Threat Intelligence (CTI) frameworks, such as MITRE ATT\&CK, CWE, CVE, CAPEC, and more. We created a dataset designed to assess models’ capabilities in understanding these interconnections. Specifically, we evaluate the model’s proficiency in mapping from tactics, techniques, attack patterns, weaknesses, and vulnerabilities to potential detections and mitigations. The ability to accurately map and explain these detections and mitigations is crucial for the model to perform effectively as a CTI security assistant. 

\setlength{\parindent}{2em}\textbf{CWE Technical Impact Mapping} In CWE, each weakness, if successfully exploited, can lead to one or more technical impacts out of eight options: modify data, read data, DoS: unreliable execution, DoS: resource consumption, execute unauthorized code or commands, gain privileges / assume identity, bypass protection mechanism, and hide activities. We have developed an evaluation set that presents the model with CWEs and their descriptions, and the goal is to classify each CWE to its related technical impact. To ensure an accurate evaluation, we selected only CWEs with exactly one impact. For CWEs with multiple possible impacts, we converted the question into a multiple-choice format.

\setlength{\parindent}{2em}\textbf{CISSP Assessment Questions}
The Certified Information Systems Security Professional (CISSP) is a well-known recognized certification in the field of cyber-security. It validates a professional's deep technical and managerial competence in designing, engineering, and managing an organization’s overall security posture. We have developed an evaluation set based on multiple-choice questions drawn from the assessment tests within the CISSP learning material.

\setlength{\parindent}{2em}\textbf{MMLU Computer Security (SecMMLU)}
Is a subset of the MMLU (Measuring Massive Multitask Language Understanding) \cite{hendrycks2020measuring} specifically focused on the computer security domain. The original MMLU encompasses multiple-choice questions from a variety of fields, but only those about computer security are utilized in our evaluation.

\setlength{\parindent}{2em}\textbf{Cybersecurity Skill Assessment} Is a multiple-choice cyber-security subset of the practice questions used for professional skill assessments on LinkedIn. These questions aim to evaluate candidates' general knowledge in the cyber-security domain 
\footnote{https://github.com/Ebazhanov/linkedin-skill-assessments-quizzes}.

\setlength{\parindent}{2em}\textbf{CyberMetric}
CyberMetric \cite{tihanyi2024cybermetric} is a benchmark dataset for evaluating Large Language Models Knowledge in  cyber-security.
The questions for the benchmark were created through a collaborative process, i.e., merging expert knowledge with LLMs. We used the 500-question dataset, verified by human evaluators, which covers a wide range of topics within cyber-security, chosen by 30 security experts.

\setlength{\parindent}{2em}\textbf{Cyber Threat Intelligence Multiple Choice Questions (CTI-MCQ)}
CTI-MCQ is a benchmark dataset developed by \cite{alam2024ctibench} for assessing LLMs' knowledge and capabilities on attack patterns, threat actors, APT campaigns, detection methods, mitigation strategies, common software vulnerabilities, attack pattern enumeration, alongside public CTI quizzes. 

\setlength{\parindent}{2em}\textbf{SecEval}
SecEval \cite{li2023seceval} is a benchmark for evaluating cyber-security knowledge in Foundation Models (FMs), offering over 2000 multi-choice, multi-option questions across 9 domains, generated by OpenAI GPT-4 using authoritative sources such as open-licensed textbooks, official documentation, and industry guidelines and standards.

\myparagraph{Classification Tasks}

\setlength{\parindent}{2em}\textbf{CTI Relationship Prediction}
A major role of our model is to learn and understand the relationships between different CTI frameworks. For example, it must determine if and how a given CVE and CWE are related. To test this ability, we have built a dataset that presents the model with two entities (e.g., instances of CVE and CWE) and two possible explanations—one explaining why the entities are related and another explaining why they are not. The model’s objective is to classify which explanation is correct, or, in other words, to determine if the two entities are related or not.

\setlength{\parindent}{2em}\textbf{CTI Entity Classification}
We have developed a dataset consisting of various descriptions corresponding to different CTI entities (such as tactics, techniques, software, etc.). The model's objective is to classify whether a given description is related to the specified entity. 

\setlength{\parindent}{2em}\textbf{Cyber Threat Intelligence Root Cause Mapping (CTI-RCM)}
CTI-RCM was developed by \cite{alam2024ctibench} to identify the fundamental causes of vulnerabilities by correlating CVE records and bug tickets with their associated weaknesses (CWE entities). Accurate root cause mapping is essential for guiding investments, policies, and practices aimed at addressing and eliminating these vulnerabilities. 

\myparagraph{Summarization Tasks}

\setlength{\parindent}{2em}\textbf{CWE Description Summarization}
We have developed a dataset containing weaknesses from the CWE dataset, intending to summarize the extended descriptions of each CWE. The target of the summarization is the short description provided for each CWE, which aims to offer a concise explanation of the CWE’s extended description.



\section{Experiments} 
\label{results}

\subsection{CyberPal.AI training details}
Similar to \citep{mitra2023orca}, we've empirically noticed that presenting the model with instructions of increasing length improves the model's learning ability. We extend on this idea and employ an incremental training methodology, organized at the dataset level. We sort the datasets of SecKnowledge into two hierarchical orders: first, we sequence the datasets by their data source category, with instructions from simpler data sources introduced first. For example - BRON-related instructions will be presented only after we present the model with MITRE ATT\&CK and the other frameworks BRON is composed of. In the second hierarchy, within each category, we arrange the instructions based on the increasing length of their outputs.

To train CyberPal.AI, we use our generated \textit{SecKnowledge} instruction dataset.
As our baseline models,  we used Llama-3 instruct 8B \citep{llama3modelcard}, Mistral instruct 7B v0.3 \citep{jiang2023mistral}, and Phi-3-medium-4k-instruct \citep{abdin2024phi}.

We employ a learning rate of $4e-5$ for Llama and Phi, and $3e-5$ for mistral. Additionally, we employ linear warm-up for 125 steps. The context length is set to 4096, and an effective batch size of 2048 is achieved using gradient accumulation.
Based on our empirical findings, beyond 2 epochs, we observed that additional epochs have negligible impact on the final loss before the model starts to overfit.

\subsection{Evaluation metrics}
Assessing LLMs on selected datasets requires appropriate evaluation metrics. We apply suitable metrics for each task as described below.
For Multiple-choice Q\&A, we employ the common HELM \cite{liang2022holistic} evaluation method where the token with maximum probability is chosen. 
For summarization tasks, we use the Recall-Oriented Understudy for Gisting Evaluation (ROUGE) \cite{lin2004rouge}.
Finally, for classification tasks, we use accuracy as the metric.
To calculate the average score for evaluation tasks, a straightforward averaging technique is utilized. For summarization tasks specifically, the mean of ROUGE-1, ROUGE-2, and ROUGE-L scores is first determined before calculating the overall average. All evaluation tasks were done in a zero-shot setting.

\begin{table}[h]
\scalebox{0.73}{
\renewcommand{\arraystretch}{1.5} 
\noindent\hspace*{-1.4cm} 
\begin{tabular}{l|cccccccc}
\hline
Model & \multicolumn{1}{l}{\begin{tabular}[c]{@{}l@{}}Original/Adv. \\ MITRE \\ ATTACK\end{tabular}} & \multicolumn{1}{l}{\begin{tabular}[c]{@{}l@{}}SIEM Rule \\ TTP Mapping\end{tabular}} & \multicolumn{1}{l}{\begin{tabular}[c]{@{}l@{}}CTI Detection \\ and Mitigation\end{tabular}} & \multicolumn{1}{l}{\begin{tabular}[c]{@{}l@{}}CWE \\ Summarization\\ (R-1/2/L)\end{tabular}} & \multicolumn{1}{l}{\begin{tabular}[c]{@{}l@{}}Technical \\ Impact \\ Mapping\end{tabular}} & \multicolumn{1}{l}{\begin{tabular}[c]{@{}l@{}}CTI \\ Relationship\\ Prediction\end{tabular}} & \multicolumn{1}{l}{\begin{tabular}[c]{@{}l@{}}CTI \\ Entity \\ Classification\end{tabular}} & Avg. \\ \hline
Mistral-7B-Instruct-v0.3 & 73.24/59.57 & 52.05 & 56.22 & 28.25/8.16/20.57 & 59.59 & 52.44 & 65.31 & 52.02 \\
\textbf{CyberPal.AI-Mistral} & \textbf{98.87/92.54} & \textbf{67.12} & \textbf{70.26} & \textbf{56.78/51.79/54.71} & \textbf{69.05} & \textbf{97.81} & \textbf{83.66} & \textbf{76.41} \\ \hline
Meta-Llama-3-8B-Instruct & 78.59/59.57 & 60.27 & 55.77 & 26.38/8.16/18.33 & 59.02 & 59.76 & 55.77 & 52.54 \\
\textbf{CyberPal.AI-Llama} & \textbf{97.04/87.74} & \textbf{64.38} & \textbf{81.95} & \textbf{46.43/38.45/43.88} & \textbf{66.18} & \textbf{97.81} & \textbf{81.95} & \textbf{74.70} \\ \hline
Phi-3-medium-4k-instruct & 77.32/64.50 & 55.47 & 67.92 & 27.96/7.83/19.94 & 66.76 & 63.75 & 67.92 & 57.84 \\
\textbf{CyberPal.AI-Phi} & \textbf{96.76/89.57} & \textbf{65.07} & \textbf{81.24} & \textbf{48.23/39.24/44.67} & \textbf{68.20} & \textbf{96.27} & \textbf{81.24} & \textbf{75.10} \\ \hline
\end{tabular}%
}
\caption{Evaluation results for CyberPal.AI models compared to the base model on designated datasets constructed to evaluate the models' performance on training-aligned security tasks. For the MITRE ATT\&CK evaluation dataset, we provide results for both the original evaluation dataset and its adversarial version, where we can see that our fine-tuned versions demonstrate greater robustness.}
\label{tab:results1}
\end{table}

\begin{table}[h]
\centering
\scalebox{0.8}{
\renewcommand{\arraystretch}{1.5} 
\noindent\hspace*{-1.2cm} 
\begin{tabular}{l|cccccccc}
\hline
Model & \multicolumn{1}{l}{\begin{tabular}[c]{@{}l@{}}CISSP \\ Assessment\end{tabular}} & \multicolumn{1}{l}{SecMMLU} & \multicolumn{1}{l}{\begin{tabular}[c]{@{}l@{}}Cybersecurity \\ Skill Assessment\end{tabular}} & \multicolumn{1}{l}{CyberMetric} & CTI-MCQ & CTI-RCM & SecEval & Avg. \\ \hline
Mistral-7B-Instruct-v0.3 & 63.63 & 67.00 & \textbf{78.69} & 80.80 & 58.03 & 45.85 & 32.98 & 60.99 \\
\textbf{CyberPal.AI-Mistral} & \textbf{89.93} & \textbf{74.00} & 78.11 & \textbf{81.60} & \textbf{65.33} & \textbf{58.20} & \textbf{42.30} & \textbf{69.92} \\ \hline
Meta-Llama-3-8B-Instruct & 71.71 & 74.00 & 82.24 & 83.20 & 63.28 & 41.45 & 32.61 & 64.07 \\
\textbf{CyberPal.AI-Llama} & \textbf{90.40} & \textbf{77.00} & \textbf{86.98} & \textbf{84.80} & \textbf{66.41} & \textbf{60.65} & \textbf{55.04} & \textbf{74.47} \\ \hline
Phi-3-medium-4k-instruct & 77.27 & 78.00 & 83.43 & 87.20 & 65.53 & 30.68 & 45.36 & 66.78 \\
\textbf{CyberPal.AI--Phi} & \textbf{90.40} & \textbf{80.00} & \textbf{86.39} & \textbf{91.00} & \textbf{72.65} & \textbf{53.00} & \textbf{67.47} & \textbf{77.27} \\ \hline
\end{tabular}%
}
\caption{Evaluation results for CyberPal.AI models compared to the base on public and general cyber-security benchmarks datasets. Although our fine-tuned models were not trained on these types of tasks, they exhibit significant and consistent improvement over baselines.}
\label{tab:results2}
\end{table}

\subsection{CyberPal.AI Results}
To demonstrate the effectiveness of \textit{SecKnowledge}, we present evaluation results of SecKnowledge-Eval for both baseline models and their fine-tuned version, trained using \textit{SecKnowledge}. As mentioned above, we used Llama-3 instruct 8B, Mistral instruct 7B v0.3, and Phi-3-medium-4k-instruct (a 13B parameters model) as our baseline models. We tested each model before and after applying fine-tuning with \textit{SecKnowledge}. Results  are presented in Tables \ref{tab:results1} and \ref{tab:results2}.  Additional results for Gemma-2 models \citep{gemma_2024} can be found in Appendix \ref{additinal_results}.

As can be seen by Table \ref{tab:results1}, on training aligned tasks (i.e., tasks that the model has seen during fine-tuning), our fine-tuned CyberPal.AI models exhibit significant and consistent improvement across different tasks, which include Multi-choice Q\&A, Summarization, and classification.
\textbf{\textit{Overall, our fine-tuned versions obtained a significant average improvement of 18-24\% across all CTI evaluation datasets.}}

As a notable instance, when testing the model with the Adversarial MITRE evaluation dataset and its non-adversarial version, our fine-tuned CyberPal.AI models demonstrate greater robustness compared to the base models. This finding suggests that CyberPal.AI is more resilient and has successfully generalized to the domain of cyber-security.
More specifically, for Mistral, our fine-tuned model exhibits a degradation of 6\% in accuracy when tested with the adversarial version of the dataset, compared to Mistral, which exhibits a degradation of 14\%.
The same goes for Llama, where our fine-tuned model exhibits a degradation of 9\% in accuracy 
with the adversarial version of the dataset, compared to Llama, which shows a degradation of 19\%.
Lastly, for Phi, our fine-tuned model exhibits a degradation of 7\% in accuracy 
with the adversarial version of the dataset, compared to Phi which shows a degradation of 13\%.
Overall, these results demonstrate the robust knowledge CyberPal.AI gained during our fine-tuning process. See  Appendix \ref{adv_mc_appendix} for more details.

Additionally, we compared the performance of CyberPal.AI with seven general and public cyber-security benchmarks to test if our fine-tuned models could improve on different types of public cyber-security benchmarks.
In Table \ref{tab:results2}, we tested seven general and public cyber-security evaluation datasets: the cyber-security section in MMLU (SecMMLU), LinkedIn cyber-security skills assessment test, publicly available CTI evaluation dataset, CTI root cause mapping dataset, and two additional cyber-security benchmarks named CyberMetric and SecEval. We also extracted multiple-choice Q\&A questions from the CISSP assessment tests, aimed at validating security analysts’ understanding of cyber-security.
\textbf{\textit{Overall, our fine-tuned models achieved an average improvement of 9-10\% across the general cyber-security evaluation datasets.}} This is impressive, considering that our models were fine-tuned with different kinds of datasets, which also indicates that our fine-tuned models managed to generalize and better understand the domain.

In summary, our evaluation of CyberPal.AI models on diverse data sources, including both proprietary and public datasets, demonstrated significant and consistent enhancements across multiple tasks, such as multiple-choice Q\&A, classification, and summarization.

\section{Conclusion}
In this work, we introduced SecKnowledge, SecKnowledge-Eval, and CyberPal.AI. 
SecKnowledge is a domain-knowledge-driven cyber-security instruction dataset aimed at fine-tuning LLMs for the security domain. The dataset construction involves two main steps. In the first step, we create instructions based on predefined schemas established through domain expertise. In the second step, we expand the initial dataset through a hybrid synthetic content-grounded data generation process.
CyberPal.AI represents a family of LLMs fine-tuned using SecKnowledge, aimed at developing security-specialized models capable of answering and following complex security-related instructions. 
To evaluate CyberPal.AI, we introduced SecKnowledge-Eval, a comprehensive suite of evaluation datasets that includes a diverse range of cyber-security tasks we developed, along with other publicly available security benchmarks. This suite is specifically designed to assess the performance of LLMs in the cyber-security domain.
Our fine-tuned CyberPal.AI models demonstrated impressive performance on various security-related tasks, including threat hunting (e.g., up to 26\% improvement on CTI Detection
and Mitigation), TTP mapping (e.g., up to 17\% improvement in SIEM Rule TTP mapping), summarization (e.g., up to 35\% improvement in CWE Summarization), and impact mapping (e.g., up to 11\% improvement in CWE technical impact mapping). 
Additionally, our models also effectively captured relationships between different components and concepts within various security frameworks (e.g., up to 45\% in CTI relationship prediction). CyberPal.AI also demonstrated enhanced performance on general security knowledge benchmarks such as the security portion of the MMLU, skill assessment tests, and analysts' assessment tests, among others. 
Overall, CyberPal.AI models outperformed their baseline counterparts, achieving significant average improvement of up to 24\% on training-aligned tasks and up to 10\% average improvement on public cyber-security benchmarks. These results underscore the extensive knowledge and deep understanding gained through fine-tuning the models with our SecKnowledge dataset.

\bibliography{cyberpal}

\begin{thebibliography}{46}
\providecommand{\natexlab}[1]{#1}
\providecommand{\url}[1]{\texttt{#1}}
\expandafter\ifx\csname urlstyle\endcsname\relax
  \providecommand{\doi}[1]{doi: #1}\else
  \providecommand{\doi}{doi: \begingroup \urlstyle{rm}\Url}\fi

\bibitem[Abdin et~al.(2024)Abdin, Jacobs, Awan, Aneja, Awadallah, Awadalla,
  Bach, Bahree, Bakhtiari, Behl, et~al.]{abdin2024phi}
Marah Abdin, Sam~Ade Jacobs, Ammar~Ahmad Awan, Jyoti Aneja, Ahmed Awadallah,
  Hany Awadalla, Nguyen Bach, Amit Bahree, Arash Bakhtiari, Harkirat Behl,
  et~al.
\newblock Phi-3 technical report: A highly capable language model locally on
  your phone.
\newblock \emph{arXiv preprint arXiv:2404.14219}, 2024.

\bibitem[Aghaei et~al.(2022)Aghaei, Niu, Shadid, and
  Al-Shaer]{aghaei2022securebert}
Ehsan Aghaei, Xi~Niu, Waseem Shadid, and Ehab Al-Shaer.
\newblock Securebert: A domain-specific language model for cybersecurity.
\newblock In \emph{International Conference on Security and Privacy in
  Communication Systems}, pp.\  39--56. Springer, 2022.

\bibitem[AI@Meta(2024)]{llama3modelcard}
AI@Meta.
\newblock Llama 3 model card.
\newblock 2024.
\newblock URL
  \url{https://github.com/meta-llama/llama3/blob/main/MODEL_CARD.md}.

\bibitem[Alam et~al.(2024)Alam, Bhushl, Nguyen, and Rastogi]{alam2024ctibench}
Md~Tanvirul Alam, Dipkamal Bhushl, Le~Nguyen, and Nidhi Rastogi.
\newblock Ctibench: A benchmark for evaluating llms in cyber threat
  intelligence.
\newblock \emph{arXiv preprint arXiv:2406.07599}, 2024.

\bibitem[Bayer et~al.(2022)Bayer, Kuehn, Shanehsaz, and
  Reuter]{bayer2022cysecbert}
Markus Bayer, Philipp Kuehn, Ramin Shanehsaz, and Christian Reuter.
\newblock Cysecbert: A domain-adapted language model for the cybersecurity
  domain.
\newblock \emph{arXiv preprint arXiv:2212.02974}, 2022.

\bibitem[Brown et~al.(2020)Brown, Mann, Ryder, Subbiah, Kaplan, Dhariwal,
  Neelakantan, Shyam, Sastry, Askell, et~al.]{brown2020language}
Tom Brown, Benjamin Mann, Nick Ryder, Melanie Subbiah, Jared~D Kaplan, Prafulla
  Dhariwal, Arvind Neelakantan, Pranav Shyam, Girish Sastry, Amanda Askell,
  et~al.
\newblock Language models are few-shot learners.
\newblock \emph{Advances in neural information processing systems},
  33:\penalty0 1877--1901, 2020.

\bibitem[Carlini \& Wagner(2017)Carlini and Wagner]{carlini2017adversarial}
Nicholas Carlini and David Wagner.
\newblock Adversarial examples are not easily detected: Bypassing ten detection
  methods.
\newblock In \emph{Proceedings of the 10th ACM workshop on artificial
  intelligence and security}, pp.\  3--14, 2017.

\bibitem[Chaudhary(2023)]{codealpaca}
Sahil Chaudhary.
\newblock Code alpaca: An instruction-following llama model for code
  generation.
\newblock \url{https://github.com/sahil280114/codealpaca}, 2023.

\bibitem[Chiang et~al.(2023)Chiang, Li, Lin, Sheng, Wu, Zhang, Zheng, Zhuang,
  Zhuang, Gonzalez, et~al.]{chiang2023vicuna}
Wei-Lin Chiang, Zhuohan Li, Zi~Lin, Ying Sheng, Zhanghao Wu, Hao Zhang, Lianmin
  Zheng, Siyuan Zhuang, Yonghao Zhuang, Joseph~E Gonzalez, et~al.
\newblock Vicuna: An open-source chatbot impressing gpt-4 with 90\%* chatgpt
  quality.
\newblock \emph{See https://vicuna. lmsys. org (accessed 14 April 2023)},
  2\penalty0 (3):\penalty0 6, 2023.

\bibitem[Chung et~al.(2022)Chung, Hou, Longpre, Zoph, Tay, Fedus, Li, Wang,
  Dehghani, Brahma, et~al.]{chung2022scaling}
Hyung~Won Chung, Le~Hou, Shayne Longpre, Barret Zoph, Yi~Tay, William Fedus,
  Yunxuan Li, Xuezhi Wang, Mostafa Dehghani, Siddhartha Brahma, et~al.
\newblock Scaling instruction-finetuned language models.
\newblock \emph{arXiv preprint arXiv:2210.11416}, 2022.

\bibitem[Ferrag et~al.(2023)Ferrag, Battah, Tihanyi, Debbah, Lestable, and
  Cordeiro]{ferrag2023securefalcon}
Mohamed~Amine Ferrag, Ammar Battah, Norbert Tihanyi, Merouane Debbah, Thierry
  Lestable, and Lucas~C Cordeiro.
\newblock Securefalcon: The next cyber reasoning system for cyber security.
\newblock \emph{arXiv preprint arXiv:2307.06616}, 2023.

\bibitem[Goodfellow et~al.(2014)Goodfellow, Shlens, and
  Szegedy]{goodfellow2014explaining}
Ian~J Goodfellow, Jonathon Shlens, and Christian Szegedy.
\newblock Explaining and harnessing adversarial examples.
\newblock \emph{arXiv preprint arXiv:1412.6572}, 2014.

\bibitem[Hemberg et~al.(2021)Hemberg, Kelly, Shlapentokh-Rothman, Reinstadler,
  Xu, Rutar, and O'Reilly]{hemberg2021linking}
Erik Hemberg, Jonathan Kelly, Michal Shlapentokh-Rothman, Bryn Reinstadler,
  Katherine Xu, Nick Rutar, and Una-May O'Reilly.
\newblock Linking threat tactics, techniques, and patterns with defensive
  weaknesses, vulnerabilities and affected platform configurations for cyber
  hunting, 2021.

\bibitem[Hendrycks et~al.(2020)Hendrycks, Burns, Basart, Zou, Mazeika, Song,
  and Steinhardt]{hendrycks2020measuring}
Dan Hendrycks, Collin Burns, Steven Basart, Andy Zou, Mantas Mazeika, Dawn
  Song, and Jacob Steinhardt.
\newblock Measuring massive multitask language understanding.
\newblock \emph{arXiv preprint arXiv:2009.03300}, 2020.

\bibitem[Jiang et~al.(2023)Jiang, Sablayrolles, Mensch, Bamford, Chaplot,
  Casas, Bressand, Lengyel, Lample, Saulnier, et~al.]{jiang2023mistral}
Albert~Q Jiang, Alexandre Sablayrolles, Arthur Mensch, Chris Bamford,
  Devendra~Singh Chaplot, Diego de~las Casas, Florian Bressand, Gianna Lengyel,
  Guillaume Lample, Lucile Saulnier, et~al.
\newblock Mistral 7b.
\newblock \emph{arXiv preprint arXiv:2310.06825}, 2023.

\bibitem[Jiang et~al.(2024)Jiang, Sablayrolles, Roux, Mensch, Savary, Bamford,
  Chaplot, Casas, Hanna, Bressand, et~al.]{jiang2024mixtral}
Albert~Q Jiang, Alexandre Sablayrolles, Antoine Roux, Arthur Mensch, Blanche
  Savary, Chris Bamford, Devendra~Singh Chaplot, Diego de~las Casas, Emma~Bou
  Hanna, Florian Bressand, et~al.
\newblock Mixtral of experts.
\newblock \emph{arXiv preprint arXiv:2401.04088}, 2024.

\bibitem[Jiao et~al.(2023)Jiao, Huang, Wang, Wang, Shi, and Tu]{jiao2023parrot}
Wenxiang Jiao, Jen-tse Huang, Wenxuan Wang, Xing Wang, Shuming Shi, and
  Zhaopeng Tu.
\newblock Parrot: Translating during chat using large language models.
\newblock \emph{arXiv preprint arXiv:2304.02426}, 2023.

\bibitem[Levi \& Kontorovich(2023)Levi and Kontorovich]{levi2023splitting}
Matan Levi and Aryeh Kontorovich.
\newblock Splitting the difference on adversarial training.
\newblock \emph{arXiv preprint arXiv:2310.02480}, 2023.

\bibitem[Li et~al.(2023)Li, Li, Guannan, Yang, and Yu]{li2023seceval}
Guancheng Li, Yifeng Li, Wang Guannan, Haoyu Yang, and Yang Yu.
\newblock Seceval: A comprehensive benchmark for evaluating cybersecurity
  knowledge of foundation models.
\newblock https://github.com/XuanwuAI/SecEval, 2023.

\bibitem[Liang et~al.(2022)Liang, Bommasani, Lee, Tsipras, Soylu, Yasunaga,
  Zhang, Narayanan, Wu, Kumar, et~al.]{liang2022holistic}
Percy Liang, Rishi Bommasani, Tony Lee, Dimitris Tsipras, Dilara Soylu,
  Michihiro Yasunaga, Yian Zhang, Deepak Narayanan, Yuhuai Wu, Ananya Kumar,
  et~al.
\newblock Holistic evaluation of language models.
\newblock \emph{arXiv preprint arXiv:2211.09110}, 2022.

\bibitem[Lin(2004)]{lin2004rouge}
Chin-Yew Lin.
\newblock Rouge: A package for automatic evaluation of summaries.
\newblock In \emph{Text summarization branches out}, pp.\  74--81, 2004.

\bibitem[Liu \& Low(2023)Liu and Low]{liu2023goat}
Tiedong Liu and Bryan Kian~Hsiang Low.
\newblock Goat: Fine-tuned llama outperforms gpt-4 on arithmetic tasks.
\newblock \emph{arXiv preprint arXiv:2305.14201}, 2023.

\bibitem[Liu et~al.()Liu, Shi, and Buford]{liucyberbench}
Zefang Liu, Jialei Shi, and John~F Buford.
\newblock Cyberbench: A multi-task benchmark for evaluating large language
  models in cybersecurity.

\bibitem[Longpre et~al.(2023)Longpre, Hou, Vu, Webson, Chung, Tay, Zhou, Le,
  Zoph, Wei, et~al.]{longpre2023flan}
Shayne Longpre, Le~Hou, Tu~Vu, Albert Webson, Hyung~Won Chung, Yi~Tay, Denny
  Zhou, Quoc~V Le, Barret Zoph, Jason Wei, et~al.
\newblock The flan collection: Designing data and methods for effective
  instruction tuning.
\newblock In \emph{International Conference on Machine Learning}, pp.\
  22631--22648. PMLR, 2023.

\bibitem[Luo et~al.(2023)Luo, Xu, Zhao, Sun, Geng, Hu, Tao, Ma, Lin, and
  Jiang]{luo2023wizardcoder}
Ziyang Luo, Can Xu, Pu~Zhao, Qingfeng Sun, Xiubo Geng, Wenxiang Hu, Chongyang
  Tao, Jing Ma, Qingwei Lin, and Daxin Jiang.
\newblock Wizardcoder: Empowering code large language models with
  evol-instruct.
\newblock \emph{arXiv preprint arXiv:2306.08568}, 2023.

\bibitem[Mitra et~al.(2023)Mitra, Del~Corro, Mahajan, Codas, Simoes, Agarwal,
  Chen, Razdaibiedina, Jones, Aggarwal, et~al.]{mitra2023orca}
Arindam Mitra, Luciano Del~Corro, Shweti Mahajan, Andres Codas, Clarisse
  Simoes, Sahaj Agarwal, Xuxi Chen, Anastasia Razdaibiedina, Erik Jones, Kriti
  Aggarwal, et~al.
\newblock Orca 2: Teaching small language models how to reason.
\newblock \emph{arXiv preprint arXiv:2311.11045}, 2023.

\bibitem[Omar \& Shiaeles(2023)Omar and Shiaeles]{omar2023vuldetect}
Marwan Omar and Stavros Shiaeles.
\newblock Vuldetect: A novel technique for detecting software vulnerabilities
  using language models.
\newblock In \emph{2023 IEEE International Conference on Cyber Security and
  Resilience (CSR)}, pp.\  105--110. IEEE, 2023.

\bibitem[Ouyang et~al.(2022)Ouyang, Wu, Jiang, Almeida, Wainwright, Mishkin,
  Zhang, Agarwal, Slama, Ray, et~al.]{ouyang2022training}
Long Ouyang, Jeffrey Wu, Xu~Jiang, Diogo Almeida, Carroll Wainwright, Pamela
  Mishkin, Chong Zhang, Sandhini Agarwal, Katarina Slama, Alex Ray, et~al.
\newblock Training language models to follow instructions with human feedback.
\newblock \emph{Advances in neural information processing systems},
  35:\penalty0 27730--27744, 2022.

\bibitem[Park \& You(2023)Park and You]{park2023pretrained}
Youngja Park and Weiqiu You.
\newblock A pretrained language model for cyber threat intelligence.
\newblock In \emph{Proceedings of the 2023 Conference on Empirical Methods in
  Natural Language Processing: Industry Track}, pp.\  113--122, 2023.

\bibitem[Raffel et~al.(2020)Raffel, Shazeer, Roberts, Lee, Narang, Matena,
  Zhou, Li, and Liu]{raffel2020exploring}
Colin Raffel, Noam Shazeer, Adam Roberts, Katherine Lee, Sharan Narang, Michael
  Matena, Yanqi Zhou, Wei Li, and Peter~J Liu.
\newblock Exploring the limits of transfer learning with a unified text-to-text
  transformer.
\newblock \emph{Journal of machine learning research}, 21\penalty0
  (140):\penalty0 1--67, 2020.

\bibitem[Ranade et~al.(2021)Ranade, Piplai, Joshi, and Finin]{ranade2021cybert}
Priyanka Ranade, Aritran Piplai, Anupam Joshi, and Tim Finin.
\newblock Cybert: Contextualized embeddings for the cybersecurity domain.
\newblock In \emph{2021 IEEE International Conference on Big Data (Big Data)},
  pp.\  3334--3342. IEEE, 2021.

\bibitem[Sanh et~al.(2021)Sanh, Webson, Raffel, Bach, Sutawika, Alyafeai,
  Chaffin, Stiegler, Scao, Raja, et~al.]{sanh2021multitask}
Victor Sanh, Albert Webson, Colin Raffel, Stephen~H Bach, Lintang Sutawika,
  Zaid Alyafeai, Antoine Chaffin, Arnaud Stiegler, Teven~Le Scao, Arun Raja,
  et~al.
\newblock Multitask prompted training enables zero-shot task generalization.
\newblock \emph{arXiv preprint arXiv:2110.08207}, 2021.

\bibitem[Sun et~al.(2024)Sun, Shen, Zhou, Zhang, Chen, Cox, Yang, and
  Gan]{sun2024principle}
Zhiqing Sun, Yikang Shen, Qinhong Zhou, Hongxin Zhang, Zhenfang Chen, David
  Cox, Yiming Yang, and Chuang Gan.
\newblock Principle-driven self-alignment of language models from scratch with
  minimal human supervision.
\newblock \emph{Advances in Neural Information Processing Systems}, 36, 2024.

\bibitem[Taori et~al.(2023)Taori, Gulrajani, Zhang, Dubois, Li, Guestrin,
  Liang, and Hashimoto]{taori2023stanford}
Rohan Taori, Ishaan Gulrajani, Tianyi Zhang, Yann Dubois, Xuechen Li, Carlos
  Guestrin, Percy Liang, and Tatsunori~B Hashimoto.
\newblock Stanford alpaca: An instruction-following llama model, 2023.

\bibitem[Team(2024)]{gemma_2024}
Gemma Team.
\newblock Gemma.
\newblock 2024.
\newblock \doi{10.34740/KAGGLE/M/3301}.
\newblock URL \url{https://www.kaggle.com/m/3301}.

\bibitem[Thawkar et~al.(2023)Thawkar, Shaker, Mullappilly, Cholakkal, Anwer,
  Khan, Laaksonen, and Khan]{thawkar2023xraygpt}
Omkar Thawkar, Abdelrahman Shaker, Sahal~Shaji Mullappilly, Hisham Cholakkal,
  Rao~Muhammad Anwer, Salman Khan, Jorma Laaksonen, and Fahad~Shahbaz Khan.
\newblock Xraygpt: Chest radiographs summarization using medical
  vision-language models.
\newblock \emph{arXiv preprint arXiv:2306.07971}, 2023.

\bibitem[Tihanyi et~al.(2024)Tihanyi, Ferrag, Jain, and
  Debbah]{tihanyi2024cybermetric}
Norbert Tihanyi, Mohamed~Amine Ferrag, Ridhi Jain, and Merouane Debbah.
\newblock Cybermetric: A benchmark dataset for evaluating large language models
  knowledge in cybersecurity.
\newblock \emph{arXiv preprint arXiv:2402.07688}, 2024.

\bibitem[Touvron et~al.(2023)Touvron, Lavril, Izacard, Martinet, Lachaux,
  Lacroix, Rozi{\`e}re, Goyal, Hambro, Azhar, et~al.]{touvron2023llama}
Hugo Touvron, Thibaut Lavril, Gautier Izacard, Xavier Martinet, Marie-Anne
  Lachaux, Timoth{\'e}e Lacroix, Baptiste Rozi{\`e}re, Naman Goyal, Eric
  Hambro, Faisal Azhar, et~al.
\newblock Llama: Open and efficient foundation language models.
\newblock \emph{arXiv preprint arXiv:2302.13971}, 2023.

\bibitem[Wang et~al.(2022)Wang, Kordi, Mishra, Liu, Smith, Khashabi, and
  Hajishirzi]{wang2022self}
Yizhong Wang, Yeganeh Kordi, Swaroop Mishra, Alisa Liu, Noah~A Smith, Daniel
  Khashabi, and Hannaneh Hajishirzi.
\newblock Self-instruct: Aligning language models with self-generated
  instructions.
\newblock \emph{arXiv preprint arXiv:2212.10560}, 2022.

\bibitem[Wei et~al.(2021)Wei, Bosma, Zhao, Guu, Yu, Lester, Du, Dai, and
  Le]{wei2021finetuned}
Jason Wei, Maarten Bosma, Vincent~Y Zhao, Kelvin Guu, Adams~Wei Yu, Brian
  Lester, Nan Du, Andrew~M Dai, and Quoc~V Le.
\newblock Finetuned language models are zero-shot learners.
\newblock \emph{arXiv preprint arXiv:2109.01652}, 2021.

\bibitem[Wei et~al.(2022)Wei, Wang, Schuurmans, Bosma, Xia, Chi, Le, Zhou,
  et~al.]{wei2022chain}
Jason Wei, Xuezhi Wang, Dale Schuurmans, Maarten Bosma, Fei Xia, Ed~Chi, Quoc~V
  Le, Denny Zhou, et~al.
\newblock Chain-of-thought prompting elicits reasoning in large language
  models.
\newblock \emph{Advances in neural information processing systems},
  35:\penalty0 24824--24837, 2022.

\bibitem[Xu et~al.(2023{\natexlab{a}})Xu, Sun, Zheng, Geng, Zhao, Feng, Tao,
  and Jiang]{xu2023wizardlm}
Can Xu, Qingfeng Sun, Kai Zheng, Xiubo Geng, Pu~Zhao, Jiazhan Feng, Chongyang
  Tao, and Daxin Jiang.
\newblock Wizardlm: Empowering large language models to follow complex
  instructions.
\newblock \emph{arXiv preprint arXiv:2304.12244}, 2023{\natexlab{a}}.

\bibitem[Xu et~al.(2023{\natexlab{b}})Xu, Guo, Duan, and McAuley]{xu2023baize}
Canwen Xu, Daya Guo, Nan Duan, and Julian McAuley.
\newblock Baize: An open-source chat model with parameter-efficient tuning on
  self-chat data.
\newblock \emph{arXiv preprint arXiv:2304.01196}, 2023{\natexlab{b}}.

\bibitem[Xu et~al.(2022)Xu, Chen, Du, Shao, Wang, Li, and
  Yang]{xu2022zeroprompt}
Hanwei Xu, Yujun Chen, Yulun Du, Nan Shao, Yanggang Wang, Haiyu Li, and Zhilin
  Yang.
\newblock Zeroprompt: scaling prompt-based pretraining to 1,000 tasks improves
  zero-shot generalization.
\newblock \emph{arXiv preprint arXiv:2201.06910}, 2022.

\bibitem[Yin et~al.(2023)Yin, Liu, Yin, Zhong, Bansal, Han, and
  Chang]{yin2023dynosaur}
Da~Yin, Xiao Liu, Fan Yin, Ming Zhong, Hritik Bansal, Jiawei Han, and Kai-Wei
  Chang.
\newblock Dynosaur: A dynamic growth paradigm for instruction-tuning data
  curation.
\newblock \emph{arXiv preprint arXiv:2305.14327}, 2023.

\bibitem[Zhang et~al.(2023)Zhang, Cui, Cai, Huang, Fang, and
  Bi]{zhang2023multi}
Yue Zhang, Leyang Cui, Deng Cai, Xinting Huang, Tao Fang, and Wei Bi.
\newblock Multi-task instruction tuning of llama for specific scenarios: A
  preliminary study on writing assistance.
\newblock \emph{arXiv preprint arXiv:2305.13225}, 2023.

\end{thebibliography}
\bibliographystyle{cyberpal}

\appendix
\section{MITRE Datasets in-depth description}
\label{mitre_data_appendix}
\paragraph{MITRE ATT\&CK:}is a comprehensive knowledge base of adversary tactics and techniques based on real-world observations. It provides a common language and structure that enables security practitioners to describe, assess, and respond cyber threats and attacks. 

More specifically, MITRE ATT\&CK consists of 14 tactics, representing the high-level goals of adversaries, such as lateral movement, initial access, and execution. Each tactic has a set of techniques, which describe the specific methods used to achieve the tactic's goals. For example, the lateral movement tactic includes techniques like internal spear-phishing. Moreover, some techniques are further divided into sub-techniques, providing even greater granularity. The ATT\&CK framework also includes information on attack campaigns, adversary groups, and the software tools they commonly use.
Lastly, the framework also provides mitigations and detections to defend against the different attack vectors.

\paragraph{CWE:}is a community-developed list of software and hardware weakness types, serving as a common language for describing security vulnerabilities.
More specifically, weaknesses usually contain relationships to other weaknesses, implementation, affected platforms, consequences, examples (which are linked to CVEs), potential mitigations, and correlation to related attack patterns (which are linked to CAPEC). As with MITRE ATT\&CK, the dataset is constructed in such a way that enables us to utilize the connection between different components of each weakness, the connection between weaknesses, and the connections to other frameworks such as CVE, MITRE ATT\&CK, and CAPEC.

\paragraph{CVE:}is a dictionary of publicly known cyber-security vulnerabilities and exposures, that aims at standardizing the way we share information regarding vulnerabilities. Each CVE typically contains several key pieces of information, including a unique identifier assigned to the vulnerability, description, affected products, severity score, connection to weaknesses (CWEs), etc.

\paragraph{CAPEC:}is a structured catalog of common attack patterns that helps users understand how adversaries exploit application weaknesses and other cyber-enabled capabilities. Attack patterns are descriptions of the common attributes and approaches employed by adversaries to exploit known weaknesses in cyber-enabled capabilities. 
Attack patterns define the challenges that an adversary may face and how they go about solving them. They are derived from the concept of design patterns applied in a destructive rather than constructive context and are generated from in-depth analysis of specific real-world exploit examples.
CAPEC provides a structured way to categorize and describe these attack patterns, including information on how they work, attack severity, the likelihood of the attack, relationships with other attack patterns, execution flow, pre-requirements, required resources, and skills, consequences of the attack, connection to CWEs, attack examples that can be linked to CVEs, and how to mitigate the attack.\footnote{https://capec.mitre.org}

\section{Adversarial Multiple-choice questions generation process}
\label{adv_mc_appendix}
To increase the difficulty of multiple-choice questions, we developed a novel adversarial attack targeting closed-domain options, where the choices are drawn from a fixed list. Here’s how it works:
\begin{enumerate}
    \item Assume a multiple-choice question with choices from a closed list of size  \textit{k}.
    \item For each of the \textit{k-1}  incorrect options, we create a new classification question. This classification question retains the original question but presents only two options: the correct choice and one of the \textit{k-1} incorrect options.
    \item We then query a language model (LLM) with each of these  \textit{k-1}  binary classification questions.
    \item From the responses, we identify the incorrect options that the model is most likely to select, given the original question and the correct answer, using the conditional loss on the incorrect option.
\end{enumerate}

\begin{figure}[h]
  \centering
  \includegraphics[width=\textwidth]{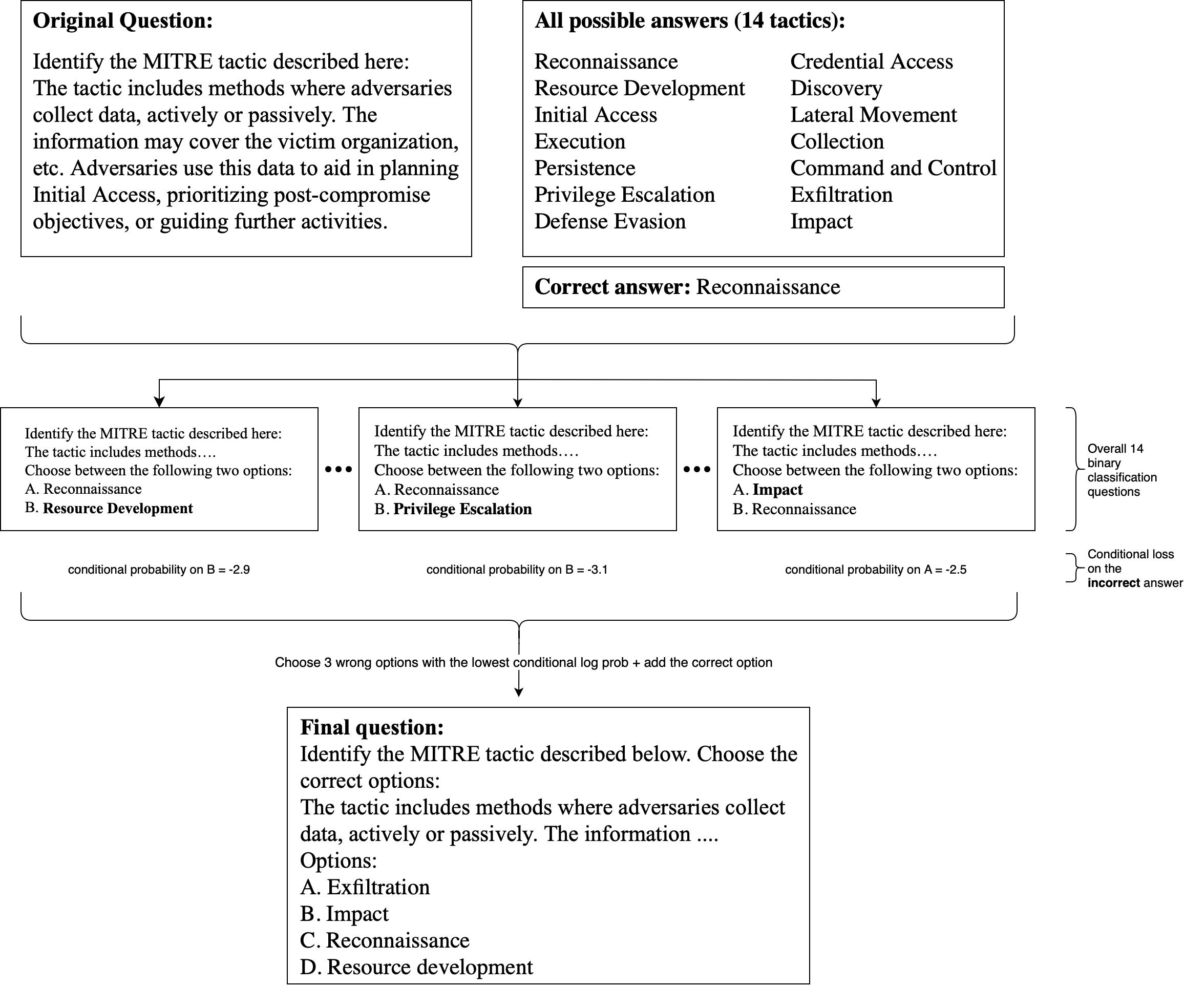}
  \caption{Adversarial MITRE ATT\&CK generation pipeline example on a question related to MITRE ATT\&CK tactics, where there are 14 possible tactics, one is the correct option, and we choose the other three options that were most likely to fool a third-party LLM.}
  \label{figures:adv-mitre}
\end{figure}

The process ensures that the false options selected are those most likely to confuse the model, thereby enhancing the overall difficulty of the dataset.
See Figure \ref{figures:adv-mitre} for an example where we asked a question related to one of the MITRE ATT\&CK tactics, and chose the other three tactics from the list of all possible tactics that are the most likely to fool a third-party LLM. The attack is an adversarial transfer attack, as we use Phi-3-small as the reference model (the model we attack), and test the attack results using the adversarial generated dataset on the other CyberPal.AI models.

Note that attacking MITRE ATT\&CK tactics requires less computational resources since the list contains only 14 possible options, but on other types of tasks, i.e., technique/software-related tasks, there are hundreds of possible options. Therefore, this attack is time and resource-consuming, but it is done only once, during the generation of the adversarial evaluation dataset.

As for the performance degradation of the models on the adversarial dataset: when testing the model with the Adversarial MITRE evaluation dataset and its non-adversarial version, our fine-tuned CyberPal.AI models demonstrate greater robustness compared to the base models. 
As can be seen in Table \ref{tab:mitre_adv_before_after}, our models experience smaller degradation in results between the adversarial and non-adversarial version of the MITRE ATT\&CK dataset: for Mistral, our fine-tuned model exhibits a degradation of 6\% in accuracy when tested with the adversarial version of the dataset, compared to the original Mistral model, which exhibits a degradation of 14\%.
The same goes for Llama, where our fine-tuned model exhibits a degradation of 9\% in accuracy when tested with the adversarial version of the dataset, compared to Llama, which shows a degradation of 19\%.
Lastly, for Phi, our fine-tuned model exhibits a degradation of 7\% in accuracy when tested with the adversarial version of the dataset, compared to Phi's base model which shows a degradation of 13\%.
These results demonstrate the robust knowledge CyberPal.AI gained during our fine-tuning process and suggest that CyberPal.AI is more resilient and has successfully
generalized to the domain of cyber-security.

\begin{table}[]
\centering
\resizebox{0.7\textwidth}{!}{%
\begin{tabular}{l|cc}
\hline
Model & \multicolumn{1}{l}{Original MITRE ATTACK} & \multicolumn{1}{l}{Adversarial MITRE ATTACK} \\ \hline
Mistral-7B-Instruct-v0.3 & 73.24 & {\color[HTML]{FE0000} 59.57 (-13.67)} \\
Sec-Mistral (Ours) & 98.87 & 92.54 (-6.3) \\ \hline
Meta-Llama-3-8B-Instruct & 78.59 & {\color[HTML]{FE0000} 59.57 (-19.0)} \\
Sec-Llama (Ours) & 97.04 & 87.74 (-9.3) \\ \hline
Phi-3-medium-4k-instruct & 77.32 & {\color[HTML]{FE0000} 64.50 (-12.8)} \\
Sec-Phi-3-medium (Ours) & 96.76 & {\color[HTML]{000000} 89.57 (-7.1)} \\ \hline
\end{tabular}%
}
\caption{Models’ results before and after applying our adversarial attack to generate the adversarial multiple-choice dataset. The “Original MITRE ATTACK” column presents the evaluation dataset results prior to the application of our adversarial method. The “Adversarial MITRE ATTACK” column shows the results after applying our adversarial technique. It is evident that CyberPal.AI models exhibit greater robustness to adversarial changes, with their results showing less drastic variation compared to those of the non-security models.}
\label{tab:mitre_adv_before_after}
\end{table}


\section{SecKnowledge-Eval statistics}
\label{eval-stat}
In Table \ref{tab:secknow-stat}, we provide high-level statistics of the different datasets composing SecKnowledge-Eval.

\begin{table}[h]
\centering
\caption{SecKnowledge-Eval statistics}
\label{tab:secknow-stat}
\begin{tabular}{l|l|c}
Eval Dataset & task type & \multicolumn{1}{r}{\# of questions} \\ \hline  
Adversarial MITRE ATT\&CK (Ours) & MCQA & 710 \\
SIEM Rule TTP Mapping (Ours) & MCQA & 146 \\
CTI Detection and Mitigation Mapping (Ours) & MCQA & 1012 \\
CWE Technical Impact Mapping (Ours) & MCQA & 349 \\
CISSP Assessment Questions (Ours) & MCQA & 206 \\
SecMMLU & MCQA & 100 \\
CyberMetric & MCQA & 500 \\
CTI-MCQ & MCQA & 2500 \\
SecEval & MCQA & 2189 \\
Cybersecurity Skill Assessment & MCQA & 170 \\
CTI-RCM & Classification & 2000 \\
CTI relationship prediction (Ours) & Classification & 778 \\
CTI Entity Classification (Ours) & Classification & 1983 \\
CWE Description Summarization (Ours) & Summarization & 92 \\
\end{tabular}
\end{table}

\section{SecKnowledge: Additional data generation details}
\subsection{BRON}
\label{bron_appendix}
In addition to the main ideas presented in the paper, we provide a more detailed explanation of the various aspects that were not included initially regarding the BRON dataset generation. We will refer to the stages demonstrated in the article.

\subparagraph{Expending BRON}
The BRON Knowledge Graph (KG) consists of the following data sources:  MITRE ATT\&CK, CAPEC, CWE, CVE, MITRE Engage, MITRE D3FEND, MITRE CAR, and, exploitdb.
We've extended BRON and added additional available information. Specifically, we've added the following information: Descriptions for technique mitigations (source: MITRE ATT\&CK), Connections between techniques and their mitigations (source: MITRE ATT\&CK), Descriptions for D3FEND mitigations (source: MITRE D3FEND), Descriptions for the relationships between Software and Technique (source: MITRE ATT\&CK), Enhanced CAPEC descriptions, including those with minimal or missing information (source: MITRE ATT\&CK). We've added the entities of the new information and managed to construct the connections between entities by using the structured nature of the MITRE data sources.

\subparagraph{Paths Extraction}
We ensured that longer paths do not include shorter ones. This is essential because we want to avoid generating similar questions and prefer unique paths to represent as many different routes as possible.

In this stage, we also gathered node pairs for negative sampling. For nodes of directly connected types, we observed that it can sometimes be quite easy to determine whether they are related, as their descriptions may be completely unrelated. To make the questions more challenging, for each node, we randomly sampled 100 nodes of the second type that are not connected to the original node. From these, we used a keywords database, and a classifier to extract features about the descriptions, to identify the node with the most similar description.

\begin{figure}[h]
  \centering
  \includegraphics[width=0.5\textwidth]{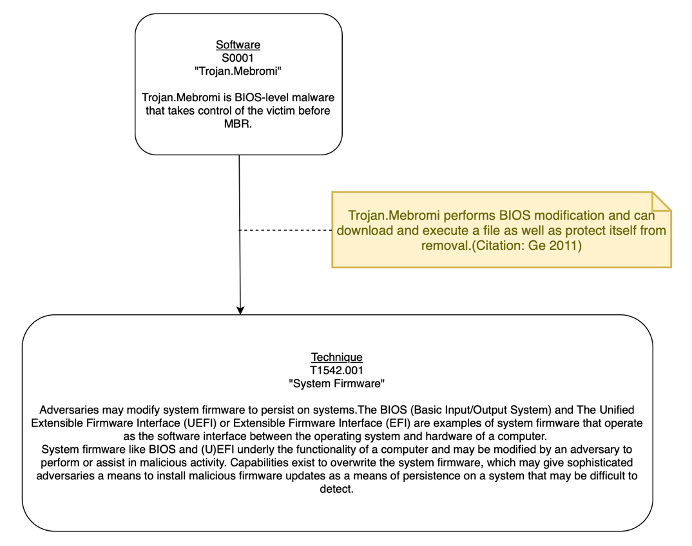}
  \caption{BRON example of existing knowledge fetched from MITRE API to explain the relation between a specific software and technique.}
  \label{figures:b1}
\end{figure}

\subparagraph{CoT on Paths}
As explained in the paper, we needed an explanation for each edge in the path to provide a complete understanding. Although the primary approach was to generate explanations using LLMs, we tried to rely as much as possible on existing knowledge. For some edge types, we had sources that explained the connections between nodes. For example, we used the MITRE API to find explanations for the connections between specific Software and Techniques (see Figure \ref{figures:b1}).

\begin{figure}[h]
  \centering
  \includegraphics[width=0.5\textwidth]{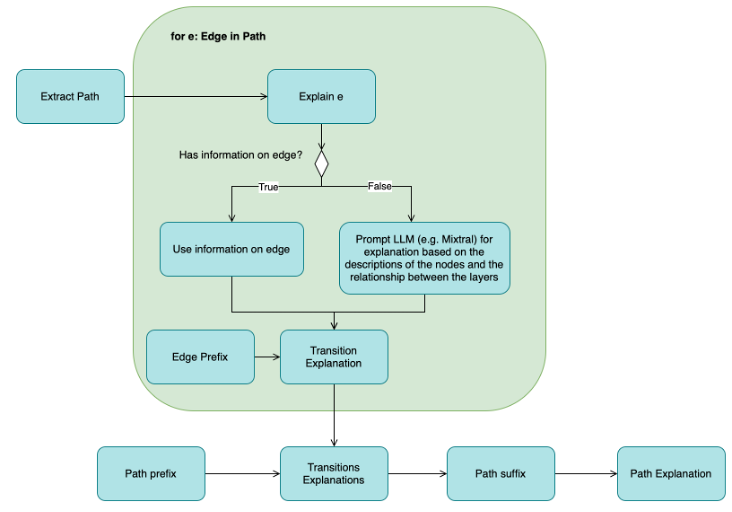}
  \caption{BRON path explanation generation process being used in explaining the selected paths. This iterative process is used heavily to construct answers that involve path explanations}
  \label{figures:b2}
\end{figure}

\subparagraph{Define Instruction schemas}
BRON can be viewed as a Directed Acyclic Graph (DAG), where nodes are categorized by type (e.g., CAPEC, CWE). Within each node type, there are direct connections (e.g., Technique to Sub-Technique) and connections to other node types (e.g., Tactic to Technique). Our goal is to enable the model to learn and understand the relationships between these different node types, which correspond to distinct datasets. See Figure \ref{figures:b2} for the generation process overview.

\begin{figure}[h]
  \centering
  \includegraphics[width=0.7\textwidth]{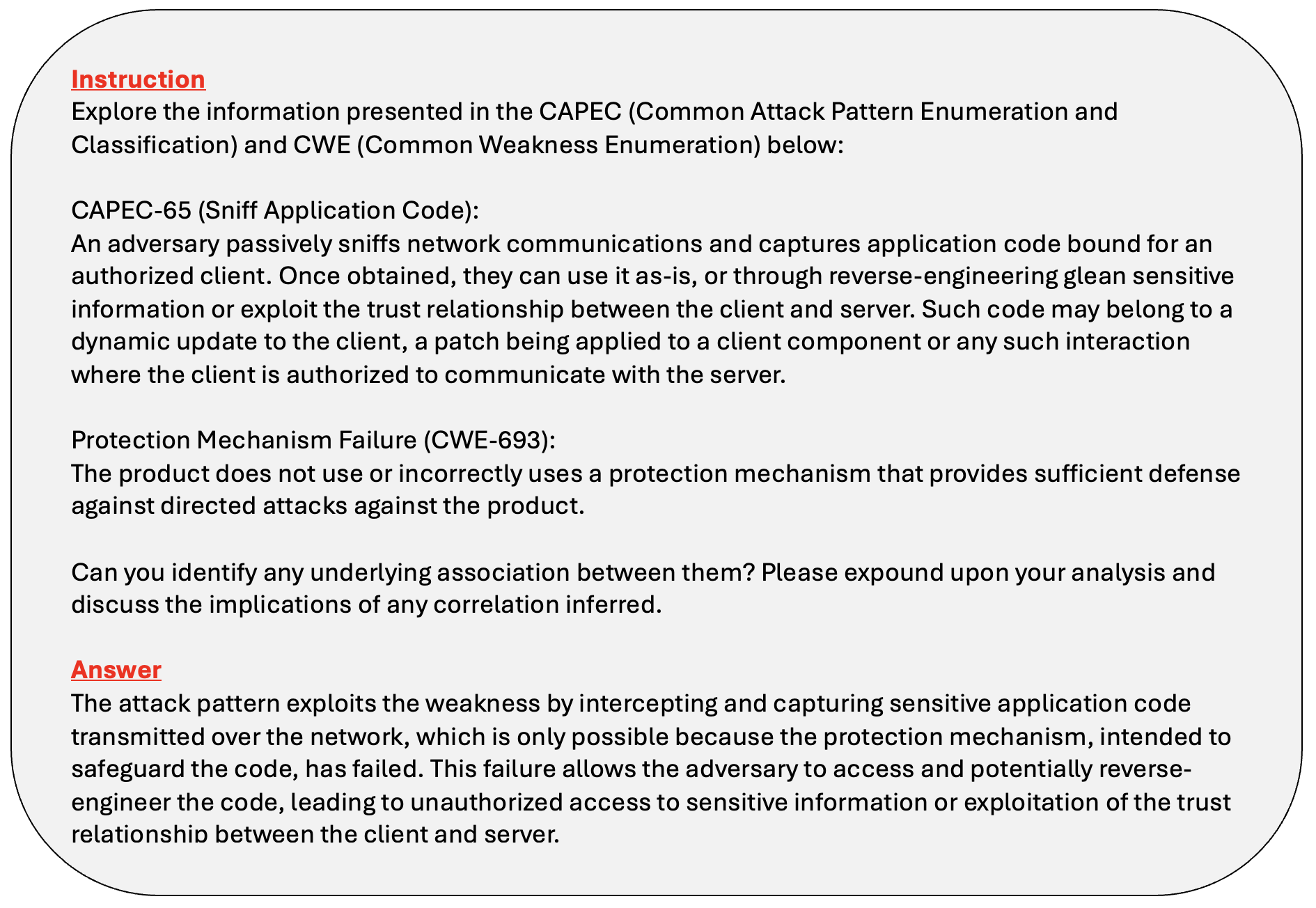}
  \caption{BRON example of direct connection instruction and its output.}
  \label{figures:b4}
\end{figure}

\begin{figure}[h]
  \centering
  \includegraphics[width=0.7\textwidth]{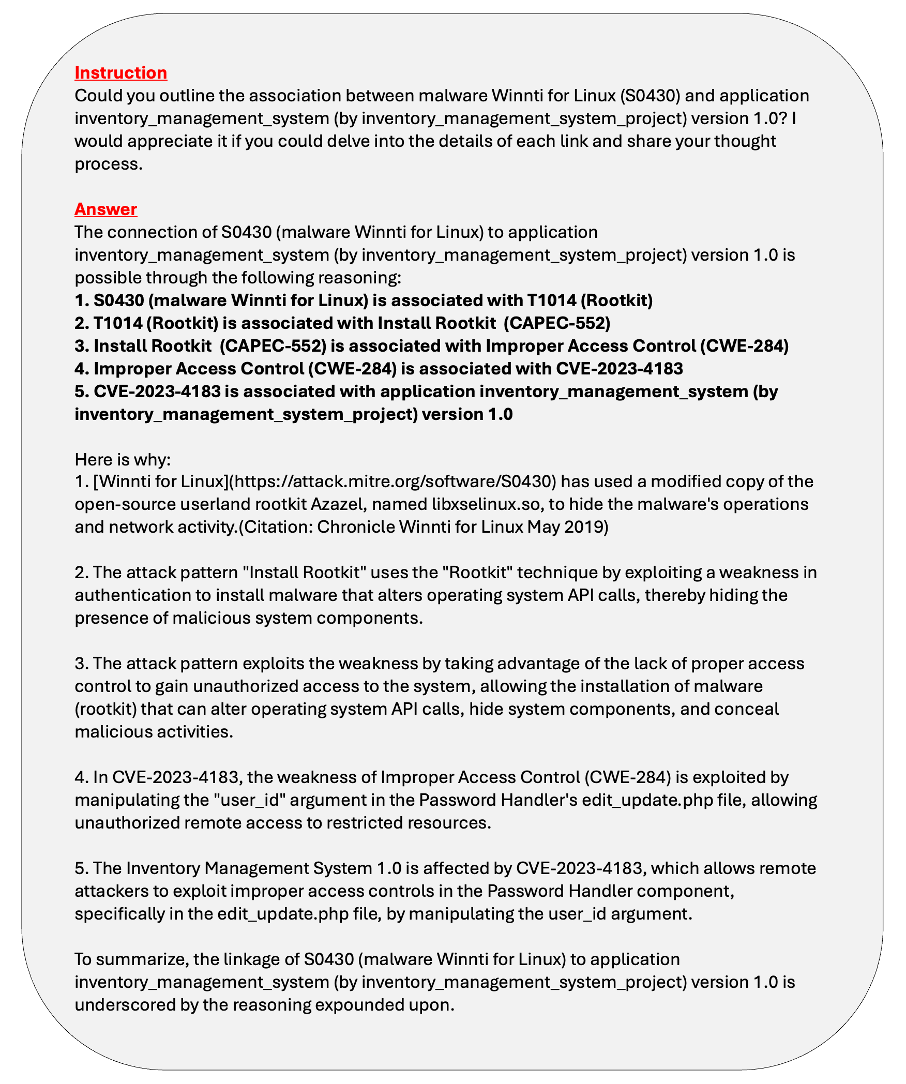}
  \caption{BRON example of node to node CoT instruction and its answer. The process used to generate the path explanation step is described in Figure \ref{figures:b2}}
  \label{figures:b3}
\end{figure}

\begin{figure}[h]
  \centering
  \includegraphics[width=0.7\textwidth]{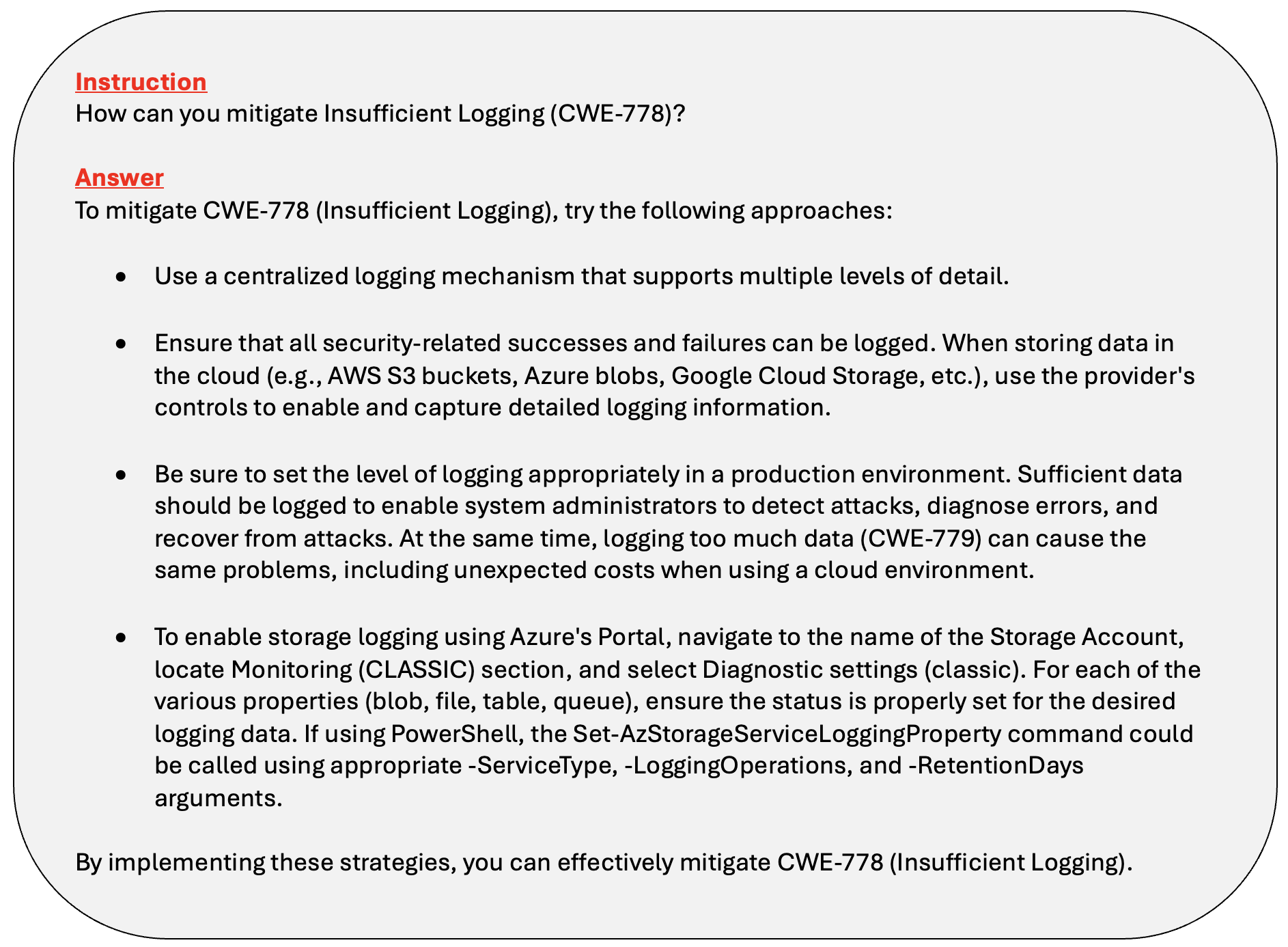}
  \caption{BRON example of direct detection/mitigation instruction and its output. Note that in this case, the data was not generated by LLM. Also, here the source node is CWE-778, and the destination node type is CWE mitigation.}
  \label{figures:b5}
\end{figure}

\begin{figure}[h]
  \centering
  \includegraphics[width=0.7\textwidth]{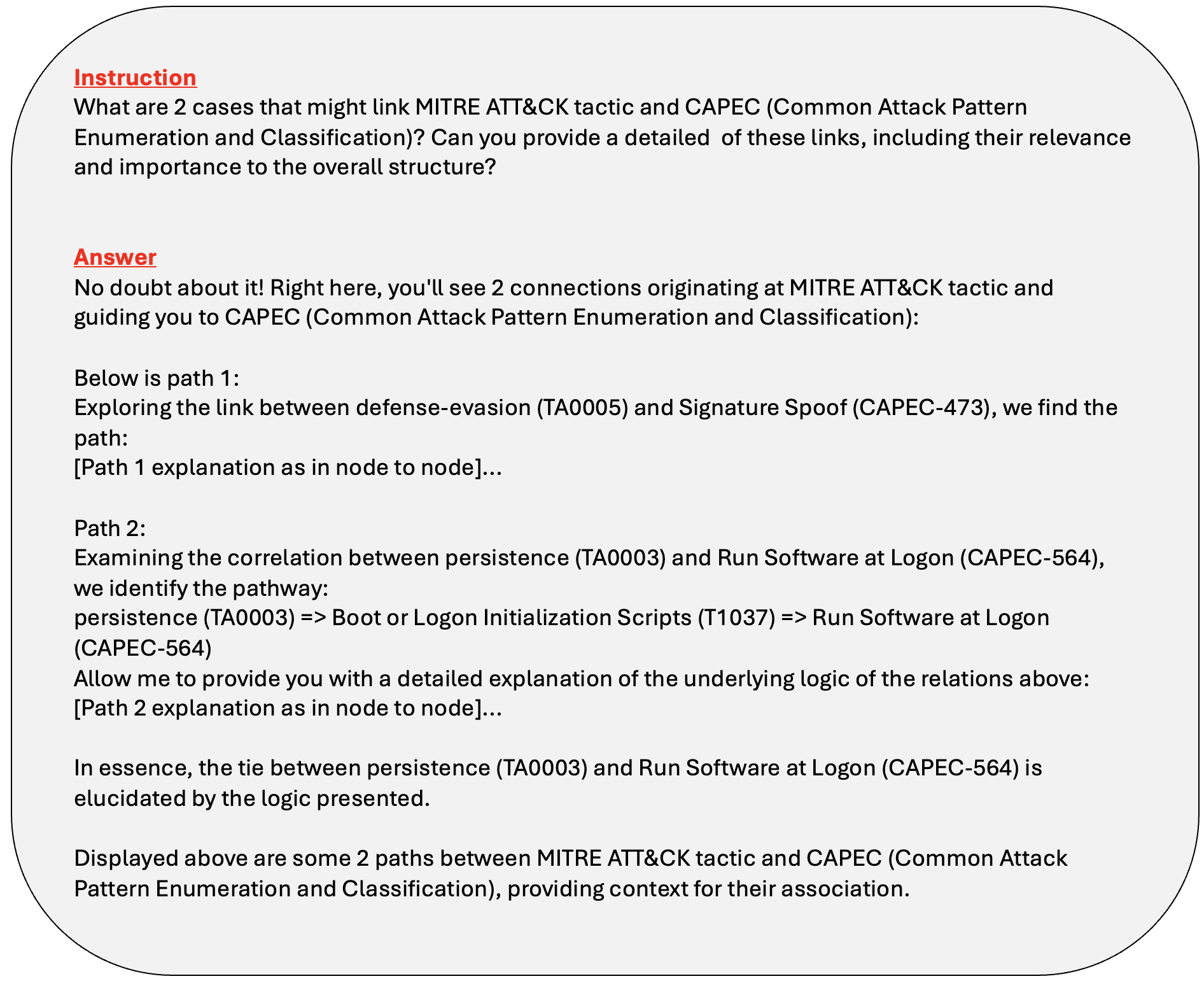}
  \caption{BRON example of node type to node type instruction and its output.}
  \label{figures:b7}
\end{figure}

After expending BRON, we define five different families of instructions:

\begin{enumerate}
  \item Direct Node to node -- Given two consecutive nodes and their descriptions, describe the relation between the nodes. See Figure \ref{figures:b4} for an example.
  \item Indirect Node to node -- Define the path/connection between 2 specific nodes. See Figure \ref{figures:b3} for an example.
  \item Node type to specific node (and vice versa) -- Describe the connection between a node type, to a specific node of another node type. See Figure \ref{figures:b5} for an example.
  \item Node type to node type -- Describe the connection between different node types using examples. See Figure \ref{figures:b7} for an example.
  \item two-step detection/mitigation -- Describe the connection between two nodes, and how to mitigate/detect the entity represented by the destination node. See \ref{fig:bron-example} for an example.
\end{enumerate}

\section{Additional Evaluation Results}
\label{additinal_results}
Here, we provide results for Google's Gemma-2 series of models. More specifically, we test the Gemma-2-2b and Gemma-2-9b models. We note that for Gemma models, specifically for Gemma-2-2b, we had to use 4-shot settings in SecEval to achieve good results.
See Tables \ref{tab:gemma-2-results} and \ref{tab:result-gemma-2-public} for results.

\begin{table}[ht]
\begin{center}
\scalebox{0.75}{
\renewcommand{\arraystretch}{1.4} 
\noindent\hspace*{-1.4cm} 
\begin{tabular}{l|ccccccc|c}
\hline
Model & \multicolumn{1}{l}{\begin{tabular}[c]{@{}l@{}}Original/Adv. \\ MITRE \\ ATTACK\end{tabular}} & \multicolumn{1}{l}{\begin{tabular}[c]{@{}l@{}}SIEM Rule \\ TTP Mapping\end{tabular}} & \multicolumn{1}{l}{\begin{tabular}[c]{@{}l@{}}CTI Detection \\ and Mitigation\end{tabular}} & \multicolumn{1}{l}{\begin{tabular}[c]{@{}l@{}}CWE \\ Summarization\\ (R-1/2/L)\end{tabular}} & \multicolumn{1}{l}{\begin{tabular}[c]{@{}l@{}}Technical \\ Impact \\ Mapping\end{tabular}} & \multicolumn{1}{l}{\begin{tabular}[c]{@{}l@{}}CTI \\ Relationship\\ Prediction\end{tabular}} & \multicolumn{1}{l|}{\begin{tabular}[c]{@{}l@{}}CTI \\ Entity \\ Classification\end{tabular}} & Avg. \\ \hline
Gemma-2-2b & 72.25/56.48 & 47.94 & 49.60 & 23.51/5.30/15.87 & 55.87 & 56.94 & 58.70 & 48.63 \\
\textbf{CyberPal.AI-Gemma-2b} & \textbf{94.64/81.97} & \textbf{60.27} & \textbf{61.66} & \textbf{35.81/43.12/44.40} & \textbf{61.32} & \textbf{95.11} & \textbf{78.62} & \textbf{68.58} \\ \hline
Gemma-2-9b & 77.46/66.62 & 47.26 & 61.46 & 23.95/6.10/16.60 & 64.18 & 70.43 & 68.28 & 56.25 \\
\textbf{CyberPal.AI-Gemma-9b} & \textbf{96.62/90.70} & \textbf{69.20} & \textbf{71.44} & \textbf{45.59/38.63/43.29} & \textbf{65.32} & \textbf{97.30} & \textbf{83.35} & \textbf{74.26} \\ \hline
\end{tabular}%
}
\caption{Evaluation results for Gemma-2 and CyberPal.AI fine-tuned models compared to the base model on designated datasets constructed to evaluate the models' performance on training-aligned security tasks. For the MITRE ATT\&CK evaluation set, we provide results for both the original evaluation set and its adversarial version, where we can see that CyberPal.AI demonstrates greater robustness.}
\label{tab:gemma-2-results}
\end{center}
\end{table}

\begin{table}[ht]
\centering
\scalebox{0.8}{
\renewcommand{\arraystretch}{1.4} 
\noindent\hspace*{-1.1cm} 
\begin{tabular}{l|ccccccc|c}
\hline
Model & \multicolumn{1}{l}{\begin{tabular}[c]{@{}l@{}}CISSP \\ Assessment\end{tabular}} & \multicolumn{1}{l}{SecMMLU} & \multicolumn{1}{l}{\begin{tabular}[c]{@{}l@{}}Cybersecurity \\ Skill Assessment\end{tabular}} & \multicolumn{1}{l}{CyberMetric} & CTI-MCQ & CTI-RCM & SecEval & Avg. \\ \hline
Gemma-2-2b & 61.62 & 65.00 & 77.51 & 77.60 & 53.79 & 3.11 & 44.90 & 54.79 \\
\textbf{CyberPal.AI-Gemma-2b} & \textbf{75.76} & \textbf{69.00} & \textbf{78.10} & \textbf{78.80} & \textbf{61.96} & \textbf{49.45} & \textbf{46.27} & \textbf{65.62} \\ \hline
Gemma-2-9b & 77.78 & 79.00 & 85.20 & 86.60 & 62.80 & 53.35 & 62.99 & 72.53 \\
\textbf{CyberPal.AI-Gemma-9b} & \textbf{89.40} & \textbf{79.00} & \textbf{85.79} & \textbf{88.00} & \textbf{68.74} & \textbf{62.80} & \textbf{64.00} & \textbf{76.82} \\ \hline
\end{tabular}%
}
\caption{Evaluation results for Gemma-2 and CyberPal.AI fine-models models compared to the base on public and general cyber-security benchmarks datasets. Although our models were not trained on these tasks, they exhibit significant and consistent improvement.
}
\label{tab:result-gemma-2-public}
\end{table}

\end{document}